\documentclass[]{ailab}

\usepackage{amsmath,amssymb}
\usepackage{array}
\usepackage{tabularx}
\newcolumntype{Y}{>{\raggedright\arraybackslash}X}
\usepackage{enumitem}
\usepackage{multirow}
\usepackage{algorithm}
\usepackage{algpseudocode}
\usepackage{float}
\usepackage{placeins}
\usepackage{needspace}
\usepackage{etoc}
\usepackage{xcolor}
\usepackage{listings}
\definecolor{skillbg}{RGB}{245,247,250}
\definecolor{skillframe}{RGB}{200,208,218}
\definecolor{promptbg}{RGB}{245,247,250}
\definecolor{promptframe}{RGB}{200,208,218}
\definecolor{skillcomment}{RGB}{110,120,135}
\definecolor{skillhead}{RGB}{176,0,64}
\definecolor{skillkey}{RGB}{0,90,160}
\lstdefinestyle{skillbox}{
  basicstyle=\footnotesize\ttfamily,
  backgroundcolor=\color{skillbg},
  frame=single,
  rulecolor=\color{skillframe},
  framesep=6pt,
  breaklines=true,
  breakindent=0pt,
  columns=fullflexible,
  keepspaces=true,
  showstringspaces=false,
  captionpos=b,
  aboveskip=6pt,
  belowskip=4pt,
  moredelim=[l][\color{skillhead}\bfseries]{\#},
  moredelim=[l][\color{skillkey}\bfseries]{name:},
  moredelim=[l][\color{skillkey}\bfseries]{description:},
  moredelim=[l][\color{skillcomment}]{--}
}
\usepackage[most]{tcolorbox}
\definecolor{cbbg}{RGB}{246,250,246}
\definecolor{cbframe}{RGB}{160,192,165}
\definecolor{cbcode}{RGB}{176,0,64}
\definecolor{cbemph}{RGB}{60,85,130}

\newcommand{\cbemphtext}[1]{{\color{cbemph}\itshape #1}}
\newtcolorbox{casebox}[1]{enhanced,breakable,sharp corners,
  colback=cbbg,colframe=cbframe,boxrule=0.7pt,fontupper=\small,
  left=8pt,right=8pt,top=1.5pt,bottom=6pt,before skip=7pt,after skip=7pt,
  before upper={\let\code\cbcodetext\let\emph\cbemphtext%
    {\bfseries\small #1}\par\nointerlineskip\vspace{2pt}%
    {\color{cbframe}\rule{\linewidth}{0.5pt}}\par\nointerlineskip\vspace{9pt}}}
\lstdefinestyle{codesnip}{basicstyle=\scriptsize\ttfamily,backgroundcolor=\color{cbbg},
  frame=single,rulecolor=\color{cbframe},framesep=5pt,columns=fullflexible,
  keepspaces=true,breaklines=true,showstringspaces=false,aboveskip=3pt,belowskip=3pt}
\definecolor{promptstr}{RGB}{0,110,90}
\definecolor{promptmark}{RGB}{150,70,0}
\lstdefinestyle{promptbox}{
  basicstyle=\scriptsize\ttfamily,
  backgroundcolor=\color{promptbg},
  frame=single,
  rulecolor=\color{promptframe},
  framesep=6pt,
  breaklines=true,
  breakindent=2em,
  columns=fullflexible,
  keepspaces=true,
  showstringspaces=false,
  aboveskip=6pt,
  belowskip=4pt,
  stringstyle=\color{promptstr},
  morestring=[b]{"},
  moredelim=[l][\color{skillhead}\bfseries]{\#},
  moredelim=[l][\color{promptmark}\bfseries]{[SYSTEM]},
  moredelim=[l][\color{promptmark}\bfseries]{[USER]},
}
\algrenewcommand{\algorithmiccomment}[1]{\hfill{\scriptsize\color{gray}\# #1}}

\newcommand{\hbsmall}{\textsf{function-as-leaf}}
\newcommand{\hblarge}{\textsf{file-as-leaf}}
\newcommand{\code}[1]{\texttt{#1}}

\title{Harness Handbook: Making Evolving Agent Harnesses Readable, Navigable, and Editable}

\author[1,2]{Ruhan Wang$^\star$}
\author[1]{Yucheng Shi$^\dagger$ }
\author[1,3]{Zongxia Li}
\author[1,4]{Zhongzhi Li}
\author[2]{Yue Yu}
\author[1,5]{Junyao Yang}
\author[1]{Kishan Panaganti}
\author[1]{Haitao Mi}
\author[2]{Dongruo Zhou}
\author[1]{Leoweiliang}

\affiliation[1]{Tencent HY LLM Frontier}
\affiliation[2]{Indiana University}
\affiliation[3]{University of Maryland, College Park}
\affiliation[4]{University of Georgia}
\affiliation[5]{National University of Singapore}
\date{July 14, 2026}
\email{ruhwang@iu.edu, yuchengshi@tencent.com}
\contribution{$^\star$Work done at an internship at Tencent Seattle $^\dagger$Lead Project Collaborator.}
\headercontent{{\sffamily\bfseries Project link:} \url{https://ruhan-wang.github.io/Harness-Handbook/}}

\abstract{
The capability of a modern AI agent depends not only on its foundation model but also on
its harness, which constructs prompts, manages state, invokes tools, and coordinates
execution. As models, APIs, execution environments, and application requirements change,
the harness must be continually modified to add capabilities or adapt existing behaviors.
Before a human developer or coding agent can make such a change, they must identify all
code locations that implement the target behavior. This is difficult because production
harnesses are often large, tightly coupled, and behaviorally distributed across files,
functions, execution stages, and state transitions, whereas modification requests describe
what the system should do and repositories are organized by files, functions, and modules.
Existing approaches to code search, repository indexing, and long-context processing make
code easier to inspect, but they still leave developers and coding agents to recover this
mapping themselves. Behavior localization is therefore a central bottleneck in harness
evolution. We introduce the \textbf{Harness Handbook}, a behavior-centric representation
synthesized automatically from a harness codebase through static program analysis and
LLM-assisted behavioral structuring, which organizes implementation knowledge around
system behaviors and links each behavior to the corresponding source code. We also
introduce \textbf{Behavior-Guided Progressive Disclosure (BGPD)}, which guides coding
agents from high-level behavior descriptions to relevant implementation details and
verifies candidate locations against the current source. We evaluate Harness Handbook on
diverse modification requests from two open-source agent harnesses. Handbook-Assisted
planning improves behavior localization and edit-plan quality while using fewer planner
tokens. The largest gains appear for changes involving scattered implementation sites,
rarely executed code paths, and cross-module interactions. These findings indicate that
evolving complex agentic systems depends not only on generating edits, but also on
determining where those edits should be made.
}

\begin{document}
\thispagestyle{firstheader}
\maketitle

\etocsettocdepth.toc{none}
\section{Introduction}

Recent advances in large language models (LLMs) have accelerated the development
of agentic systems \citep{wang2024survey,yao2022react,huang2025towards}. These
systems extend model capabilities by interacting with tools, APIs, and complex
execution environments
\citep{schick2023toolformer,zhou2024webarena,xie2024osworld,yang2024swe}. Such
interactions are not governed by the foundation model alone. They are coordinated
by the harness, which constructs prompts, manages state, invokes tools, and
controls execution across system components. The harness therefore determines
how model capabilities are translated into system behavior. As models, APIs,
execution environments, and application requirements evolve, the harness must be
adapted accordingly. Harness evolution has consequently become a recurring
engineering challenge in modern agent development
\citep{lin2026agentic,lopopolo2026harness,zhong2026ai}.

In practice, harness evolution requires changes that add capabilities, adapt
existing behaviors, or refine execution workflows
\citep{wang2025federated,wang2026fera}. Traditionally, human developers make
these changes by reading the repository, tracing the relevant behavior, and
editing its implementation. Coding agents are increasingly expected to perform
the same work from natural-language requests
\citep{yang2024swe,wang2025openhands,zhong2026ai,zhu2026semaclaw}. However, a
modification request describes what behavior should change, but it does not
identify where that behavior is implemented. Whether the modification is
performed by a human developer or a coding agent, the first step is therefore to
locate all relevant implementation sites.

Finding every relevant implementation site is difficult in production-scale
harnesses \citep{ning2026code}. A large harness may span hundreds of functions
across many files, with execution logic distributed across stages and connected
through shared state. As a result, a single behavior may depend on several
nonadjacent implementation sites.

We refer to identifying all implementation sites associated with a behavioral
request as \textbf{behavior localization}. In a large and structurally complex
harness repository, this task is costly for both human developers and coding
agents. Human developers must invest substantial time and effort in building a
mental model of the system and tracing behavior across files and execution
stages. Coding agents face an additional constraint from limited input context,
which prevents them from examining all relevant code at once. They must
therefore explore the repository iteratively and may still overlook scattered or
rarely executed paths \citep{zhang2026swe,tamoyan2026sherloc}.

Existing approaches to repository understanding make codebases easier to
explore. Repository maps, code search, code summarization, repository memory, and
long-context editing help coding agents find, inspect, and retain relevant code
\citep{zhang2026swe,bhola2026code,wang2025improving,tamoyan2026sherloc,cao2026coding}.
However, these approaches organize information around files, functions, and
modules, whereas a harness modification request describes a desired behavior.
They may identify individual pieces of relevant code, but they do not show how
those pieces work together to produce the behavior or whether every affected
location has been found. The coding agent must still connect these pieces and
infer how the requested behavior maps to the underlying implementation. This
missing connection is the gap we address.

We address this gap with \textbf{Harness Handbook}, an operational behavior
representation that connects system behavior to source code. Instead of
organizing implementation knowledge by files and functions, the Handbook
organizes it by what the harness does and links each behavior to the code that
implements it. Developers and coding agents can therefore identify the relevant
behavior first and then navigate directly to the corresponding code. We
construct Harness Handbook instances automatically using static program analysis
and LLM-assisted behavioral structuring. To evaluate whether the Handbook
improves harness modification, we use realistic requests from two open-source
agent harnesses and compare Baseline and Handbook-Assisted planning on the same
tasks. Handbook-Assisted planning improves behavior localization and edit-plan
quality while using fewer planner tokens. These results show that making the
connection between behavior and implementation explicit can make harness
evolution more accurate and efficient. Our contributions are as follows:

\begin{itemize}

\item We define \textbf{behavior localization} as finding all code locations
that implement the behavior described in a modification request. This step is
necessary before a complete edit can be planned.

\item We introduce \textbf{Harness Handbook}, which organizes implementation
knowledge by what the harness does and links each behavior directly to its
source code. We also develop an automated framework that builds Handbook
instances from existing codebases using static program analysis and LLM-assisted
behavioral structuring.

\item We introduce \textbf{Behavior-Guided Progressive Disclosure (BGPD)}, a
workflow that guides coding agents from high-level behavior descriptions to
relevant implementation details in stages.

\item We evaluate \textbf{Harness Handbook} on diverse modification requests
from two open-source harnesses. We compare Handbook-Assisted BGPD planning with
planning without Handbook access on the same tasks. Handbook assistance improves
behavior localization and edit-plan quality while using fewer planner tokens.

\end{itemize}

\section{Related Work}

Our work is closely related to three research directions: agent harnesses, harness evolution, and repository representations for coding agents. We briefly review each direction and discuss how Harness Handbook differs from existing approaches by introducing an explicit operational behavior representation for behavior localization.

\subsection{Agent Harness}

Recent work has increasingly recognized the agent harness as a first-class software abstraction that extends foundation models into deployable agentic systems. Rather than treating prompting, tool use, memory, and execution logic as implementation details, \emph{Code as Agent Harness} formalizes the harness as an executable and stateful software layer supporting reasoning, acting, feedback, and coordination \citep{ning2026code}. Complementary surveys and engineering efforts further characterize the harness as the runtime infrastructure responsible for prompt construction, state management, tool invocation, control flow, and interactions with external environments \citep{meng2026agent,he2026harness}. This abstraction has been widely adopted by modern agent frameworks and production systems. Frameworks such as AutoGen and OpenHands expose modular runtime components for agent orchestration, tool execution, environment interaction, and multi-agent collaboration \citep{wu2023autogen,wang2025openhands}. Industrial systems such as Claude Code and Codex similarly emphasize harness design as a key factor in building reliable long-running coding agents \citep{rajasekaran2026harness,openai2025codex}. While these works establish the importance of agent harnesses, they primarily focus on harness construction and system design rather than understanding and evolving existing production-scale harnesses.

\subsection{Harness Evolution}

As coding agents become increasingly capable, software engineering efforts are gradually shifting from manually implementing functionality to automatically evolving the underlying harness \citep{yang2024swe,wang2025openhands}. Recent work has begun to investigate harness engineering, adaptive harness optimization, and automated harness repair \citep{zhong2026ai,zhu2026semaclaw,lin2026agentic,chen2026harnessx,chen2026harnessforge,chen2026failed}. These approaches improve the capability and reliability of agent systems by constructing, adapting, or repairing harness implementations. Our work addresses a complementary problem. Rather than directly modifying harness implementations, we study the prerequisite challenge of harness evolution: behavior localization. Harness Handbook enables developers and coding agents to identify where behavioral changes should be made before reasoning about how to modify the underlying implementation.

\subsection{Behavior Representations}

Recent work has proposed various representations to improve repository understanding for coding agents, including structural repository representations, repository indexing, repository memory, and natural-language artifacts \citep{cherny2026repository,bhola2026code,wang2025improving,pan2026natural}. These representations improve repository navigation, code retrieval, and editing by organizing implementation knowledge into more accessible forms. However, existing representations remain fundamentally implementation-centric, organizing information around source code, repository structure, or execution infrastructure. They do not explicitly capture how system behaviors emerge across distributed execution stages, functional modules, and shared states. In contrast, Harness Handbook introduces an operational behavior representation that explicitly bridges behavioral requirements and their underlying implementation, enabling behavior localization before repository exploration.

\section{Harness Handbook}
\label{sec:overview}
Harness Handbook has three parts for understanding and modifying an agent
harness. The \emph{representation} organizes source around runtime
behavior (Section~\ref{sec:representation}). The \emph{construction} pipeline
builds this representation from a repository
(Section~\ref{sec:construction}). The \emph{modification} workflow uses the
handbook to guide code changes and automatically resynchronizes it after every
non-empty repository diff (Section~\ref{sec:modification}).

\subsection{Harness Handbook Representation}
\label{sec:representation}

\begin{figure}[t]
    \centering
    \includegraphics[width=\linewidth]{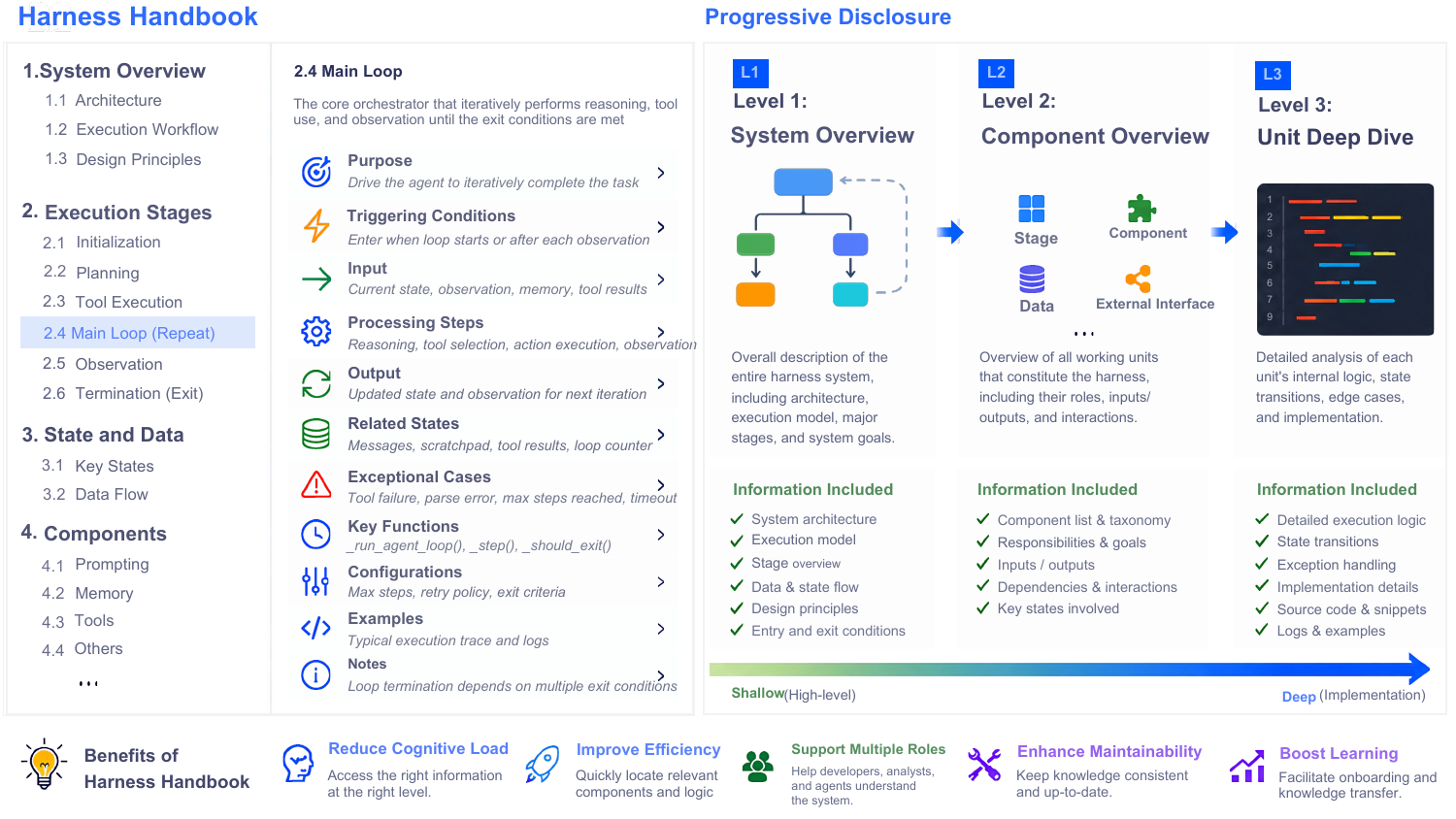}
    \caption{Overview of the Harness Handbook representation. Its three-level
    hierarchy progresses from a system-level overview to stage-level component
    overviews and source-backed unit details. The navigation pane provides
    component and state indexes for direct access and cross-stage tracing.}
    \label{fig:handbook}
\end{figure}

Source repositories show where code is stored, but they do not directly show
how a runtime behavior unfolds. One behavior may cross several files, execution
stages, and shared states. Harness Handbook reorganizes this information around
behavior while preserving links to the source. As
Figure~\ref{fig:handbook} shows, it contains an L1--L3 document tree
$\mathcal{D}$ and a complementary state-register view $\mathcal{Z}$.

A reader normally starts at \textbf{L1 (system overview)}, which summarizes the
architecture, execution model, major stages, and global data flow. The reader
then moves to \textbf{L2 (component overview)} for the responsibilities, inputs,
outputs, dependencies, and local state of a selected stage. Finally,
\textbf{L3 (unit deep dive)} links that stage to source-grounded implementation
entries. The complementary view $\mathcal{Z}$ records state relationships that
cross stage boundaries.

Two rules keep the representation useful. \emph{\textbf{Progressive
disclosure}} means that readers move from L1 to L3 only when a task requires
more detail. \emph{\textbf{Behavior--implementation alignment}} means that every
active L3 locator must still resolve to the current repository. If a locator
cannot be revalidated, its entry is frozen and excluded from localization until
it is refreshed. The repository therefore remains the authority for
implementation details.

\subsection{Harness Handbook Construction}
\label{sec:construction}

Construction builds a handbook from a repository $\mathcal{R}$. The leaf mode
$g\in\{\mathrm{function},\mathrm{file}\}$, which remains fixed for the lifetime
of the handbook, determines the granularity of L3 entries: in \hbsmall{} mode
each L3 entry covers a whole function or one or more contiguous regions, whereas
in \hblarge{} mode each L3 entry represents a file.

The two modes differ in how the stage skeleton is obtained. \hbsmall{} starts
from a seed skeleton $\mathcal{S}_0$ containing the stage and state-register
definitions; we use it when a trustworthy seed skeleton that faithfully reflects
the harness's execution stages is available and function-level L3 entries fit
the available budget. \hblarge{} instead infers the stage skeleton rather than
starting from a seed; we use it when no such seed is available or when
function-level organization would exceed the available budget. Both modes
produce the representation described in Section~\ref{sec:representation}, and
Algorithm~\ref{alg:construction} defines the two branches.

\begin{figure}[t]
    \centering
    \includegraphics[width=\linewidth]{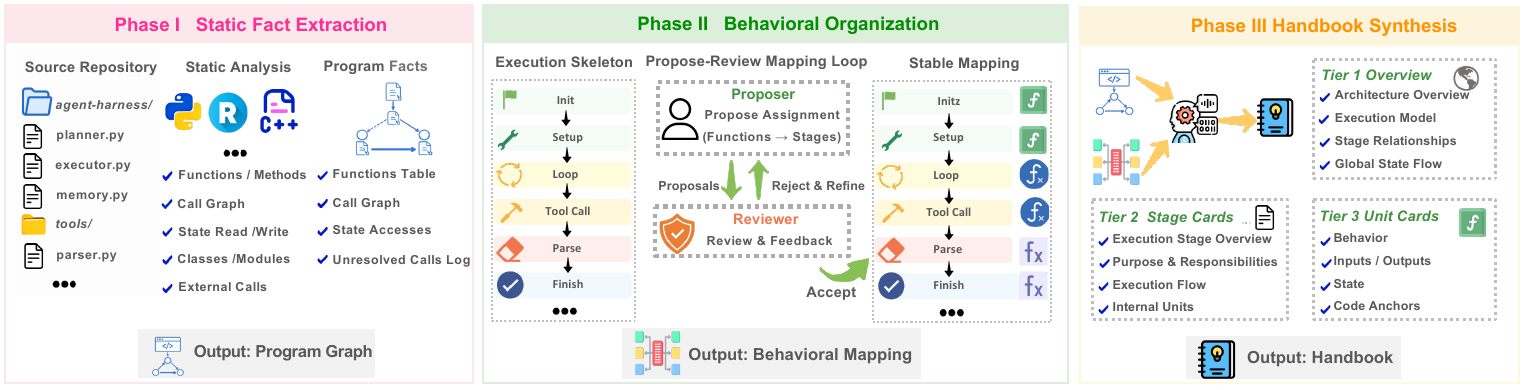}
    \caption{Construction pipeline for Harness Handbook. Static analysis extracts
    source-linked facts, behavioral organization maps source units to execution
    stages, and hierarchical synthesis builds the L1--L3 handbook.}
    \label{fig:handbook-construction}
\end{figure}

Once $g$ is chosen, construction proceeds in three phases, illustrated for the
function-as-leaf branch in Figure~\ref{fig:handbook-construction}.

\begin{enumerate}[wide=0pt,labelsep=0.5em,itemsep=4pt,topsep=3pt,label=\textbf{Phase \Roman*.}]

\item \textbf{Static Fact Extraction.} Language-specific adapters parse the
repository and extract functions, named external boundaries, source locations,
signatures, and call edges. The program graph $\mathcal{G}$ keeps only calls
that resolve to an internal function or a named boundary. Unresolved calls are
logged rather than assigned guessed targets. This phase is deterministic and
makes no LLM calls.

\item \textbf{Behavioral Organization.} The pipeline organizes source units into
the execution-stage skeleton $\mathcal{S}$ according to the selected leaf mode.
\begin{itemize}[leftmargin=1.2em,itemsep=2pt,topsep=2pt]
\item \emph{\textbf{Function-as-leaf.}} The pipeline uses source and call-graph
context to propose function-to-stage assignments, which are refined through
iterative review. A function can be assigned to one or more execution stages,
either as a whole or as contiguous regions when it serves multiple behavioral
roles. Accepted assignments and revisions update the stage skeleton
$\mathcal{S}$.
\item \emph{\textbf{File-as-leaf.}} The pipeline summarizes the scanned files as
cards and combines them with the program graph to infer the stage skeleton
$\mathcal{S}$. It then organizes the files by execution stage. An optional
proposal-and-review process iteratively refines this structure, while uncovered
files and unresolved organizational issues remain explicitly recorded in
$\mathcal{Y}$.
\end{itemize}
Appendix~\ref{app:construction-details} provides the full organization
procedures and validation checks for both modes.

\item \textbf{Hierarchical Synthesis and Packaging.} This phase converts the
stage skeleton and source organization from Phase II into the L1--L3 document
tree and cross-stage state-register view. Each L3 entry is linked to a
statically identified source location and validated against the current
repository. This keeps the handbook traceable to the source. Finally, the
pipeline renders $\mathcal{V}$ and packages the structured data needed for
source localization and future resynchronization.

\end{enumerate}

\subsection{Handbook-Guided Modification and Resynchronization}
\label{sec:modification}

After construction, the handbook serves as a behavior-oriented guide for
modifying the repository. Given a request $q$, the workflow uses the handbook
$\mathcal{H}$ together with its corresponding repository $\mathcal{R}$. The leaf
mode $g$ remains unchanged throughout the process.

The workflow consists of four steps. First, \emph{\textbf{Behavior-Guided
Progressive Disclosure (BGPD)}} localizes the requested behavior through
coarse-to-fine handbook navigation. It progressively reveals relevant
information, follows it to candidate source locations, and verifies those
locations against the current repository. This process produces a set of
source-grounded evidence for planning. Second, the planner converts the evidence
into an edit plan $\mathcal{P}$ and action declarations $\Gamma$. Third, a
separate executor applies $\mathcal{P}$ to the repository. Finally, any
non-empty diff triggers handbook resynchronization.
Algorithm~\ref{alg:bgpd} summarizes the complete workflow.

\begin{algorithm}[H]
\caption{Handbook-Guided Modification with Automatic Resynchronization}
\label{alg:bgpd}
\vspace{0.3em}
\begin{minipage}{0.94\linewidth}
\footnotesize
\begin{algorithmic}[1]
\Require request $q$, handbook package $\mathcal{H}$, corresponding repository
snapshot $\mathcal{R}$
\Ensure plan $\mathcal{P}$ and declarations $\Gamma$; updated repository
$\mathcal{R}'$, diff $\Delta$, and handbook package $\mathcal{H}'$
\State $g \leftarrow \mathrm{LeafMode}(\mathcal{H})$
\State $\widehat{\mathcal{E}}_q \leftarrow
\mathrm{BGPD}(q,\mathcal{H},\mathcal{R})$
\State $(\mathcal{P},\Gamma) \leftarrow
\mathrm{Plan}(q,\widehat{\mathcal{E}}_q)$
\State $\mathcal{R}' \leftarrow \mathrm{Execute}(\mathcal{R},\mathcal{P})$
\State $\Delta \leftarrow \mathrm{Diff}(\mathcal{R},\mathcal{R}')$
\If{$\Delta \neq \emptyset$}
    \State $\mathcal{H}' \leftarrow
    \mathrm{Resync}_{g}(\mathcal{H},\mathcal{R},\mathcal{R}',\Delta,\Gamma)$
\Else
    \State $\mathcal{H}' \leftarrow \mathcal{H}$
    \EndIf
\State \Return $(\mathcal{P},\Gamma,\mathcal{R}',\Delta,\mathcal{H}')$
\end{algorithmic}
\end{minipage}
\end{algorithm}

\subsubsection{Behavior Localization with BGPD}

Rather than treating all source locations equally, BGPD uses the handbook to
identify candidate sites from coarse to fine. It then verifies these sites
against the current repository.

Localization begins at the level of execution stages. BGPD first uses L1 and L2
of $\mathcal{D}$ to identify the stages directly relevant to the request. It
then follows $\mathcal{Z}$ to include stages coupled through shared state. This
step captures mutually dependent stages even when they are structurally distant.
Within these stages, BGPD selects the most relevant L3 entries and retrieves
their source locators.

BGPD then expands the candidate set along call relations. In \hbsmall{} mode, it
uses the function-call graph; in \hblarge{} mode, it uses the induced file-call
graph. External boundary nodes provide context but are never returned as edit
sites.

Up to this point, BGPD operates on the handbook. It then opens the current
repository $\mathcal{R}$, resolves the candidate locators, and retains only the
sites that remain relevant to $q$. This yields the verified evidence
$\widehat{\mathcal{E}}_q$. Each record contains a file path, an optional function
or region anchor, and a current source excerpt. The handbook thus guides the
search, while the repository remains the basis for the edit plan.

\subsubsection{Edit Planning and Execution}

The planner converts $\widehat{\mathcal{E}}_q$ into an edit plan $\mathcal{P}$.
Each edit block specifies where the change should be made, which source evidence
supports it, and what should change. The block records a target file, an optional
function or region anchor, a current source excerpt, and the intended change.

For each edit, the planner also records an action declaration containing the
target file, optional anchor, and one of three action types: \emph{modify},
\emph{add}, or \emph{remove}. The resulting declarations are collected as
$\Gamma=(\Gamma_{\mathrm{modify}},\Gamma_{\mathrm{add}},
\Gamma_{\mathrm{remove}})$. A rename is represented as one removal and one
addition.

The executor applies $\mathcal{P}$ to $\mathcal{R}$ and produces the updated
repository $\mathcal{R}'$. The resulting diff is
$\Delta=\mathrm{Diff}(\mathcal{R},\mathcal{R}')$. Because $\Delta$ is the
factual record of what changed, it drives handbook invalidation. By contrast,
$\Gamma$ is used only to check whether execution followed the plan.

\subsubsection{Handbook Resynchronization}

Every non-empty diff $\Delta$ triggers $\mathrm{Resync}_g$. Whenever possible,
the procedure updates only the affected parts rather than rebuilding the entire
handbook.

\textbf{First,} it reparses the changed source, refreshes the program graph, and
aligns the old and new versions to identify added, removed, and modified units:
in \hbsmall{} mode, it matches functions using fingerprints that do not depend on
line numbers, and in \hblarge{} mode, it matches files using file-set
differences and content hashes.

\textbf{Second,} it determines the update scope. If the stage skeleton
$\mathcal{S}$ remains valid, only the affected entries and their enclosing
structure are refreshed; otherwise, Algorithm~\ref{alg:construction} is rerun on
$\mathcal{R}'$ with the same leaf mode $g$ and stored configuration $\Theta$.

\textbf{Finally,} content that cannot be parsed or classified is handled
conservatively: it is frozen or recorded in the coverage record rather than
guessed.

Once the update completes, the new pair $(\mathcal{R}',\mathcal{H}')$ becomes the
starting state for the next request. Within resynchronization, model calls are
limited to four semantic steps: classification, file assignment, within-stage
organization, and description revision. All other operations are deterministic.

\FloatBarrier
\section{Experiment}
\label{sec:experiment}

We evaluate Harness Handbook on two open-source agent harnesses around three
research questions, describing the experiment setup in Section~\ref{sec:exp-setup}
and reporting results in Section~\ref{sec:exp-results}.

\begin{itemize}[leftmargin=1.4em,itemsep=1pt,topsep=1pt,parsep=0pt]
\item \textbf{RQ1:} Does handbook-guided localization improve plan quality while
reducing planning cost?
\item \textbf{RQ2:} With the handbook, can a weaker planner match the
implementation-site localization of substantially more capable models?
\item \textbf{RQ3:} Do these gains persist across request types and localization
difficulty levels?
\end{itemize}

\begin{figure}[t]
\centering
\includegraphics[width=\linewidth]{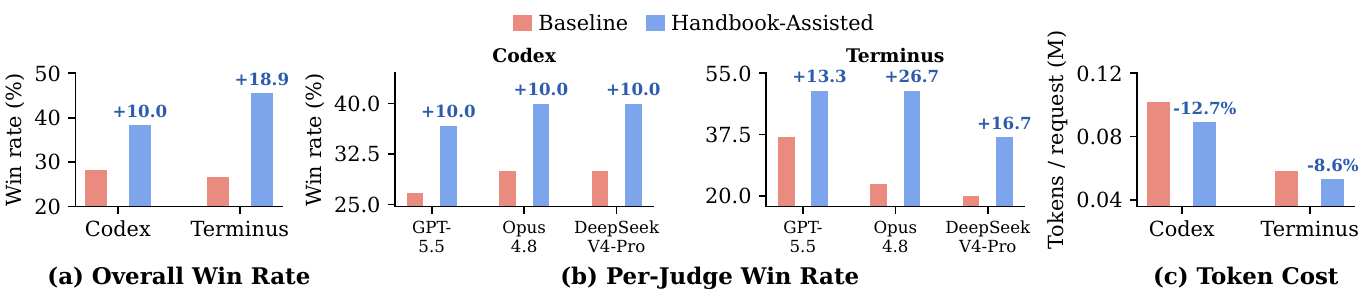}
\caption{Plan quality and planner token usage on Codex and Terminus-2.
(a) Overall win rates aggregated across the three judges. (b) Win rates reported
separately by GPT-5.5, Opus 4.8, and DeepSeek-V4-Pro. (c) Average number of
planner tokens per request; lower values indicate greater efficiency.}
\label{fig:harness-modification-results}
\end{figure}

\begin{figure}[t]
\centering
\includegraphics[width=\linewidth]{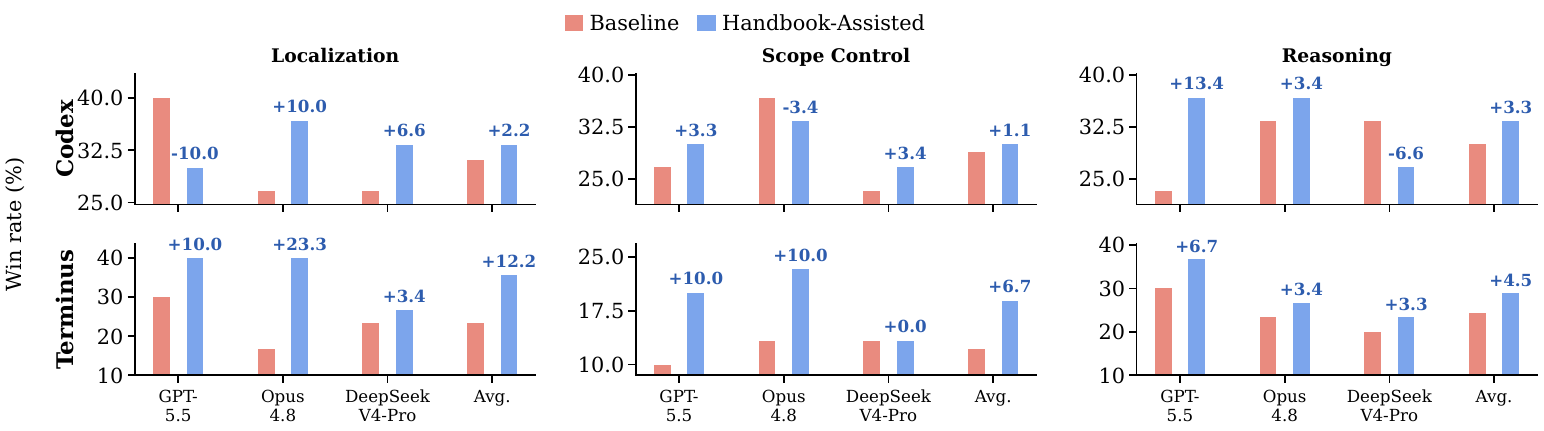}
\caption{Per-judge win rates for three evaluation dimensions. Rows show results
for Codex and Terminus-2, while columns show Localization, Scope Control, and
Reasoning.}
\label{fig:behavior-localization-results}
\end{figure}

\subsection{Experiment Setup}
\label{sec:exp-setup}

\paragraph{\textbf{Evaluation Setting.}}

This experiment evaluates localization and edit planning in the modification
workflow. Given a natural-language request, a read-only planner built with
NexAU \citep{nexagi2025nexn1} and powered by \textbf{DeepSeek-V4-Pro} produces
an edit plan $\mathcal{P}$, and we score the quality of $\mathcal{P}$ directly.

We compare two arms: the \emph{Baseline} explores the repository directly,
whereas the \emph{Handbook-Assisted} arm locates relevant source code under
handbook guidance using a BGPD-consistent navigation policy. Apart from handbook
access and navigation, the two arms are identical in requests, model,
repository, tool permissions, and decoding settings. The handbook used in the
Handbook-Assisted arm is built from the same source.

Plan quality is judged independently by three models: \textbf{GPT-5.5},
\textbf{Opus 4.8}, and \textbf{DeepSeek-V4-Pro}.

\begin{figure}[t]
\centering
\includegraphics[width=\linewidth]{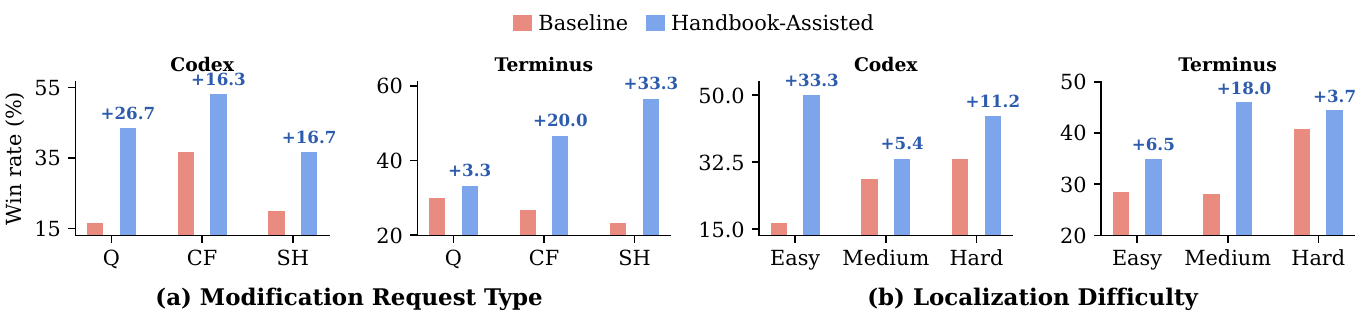}
\caption{Win rates by (a) modification request type and (b) localization
difficulty. Q, CF, and SH denote Query, Cross-file, and Search-Hostile requests.}
\label{fig:scenario-generalization-results}
\end{figure}

\paragraph{\textbf{Benchmarks and Modification Requests.}}

We evaluate \textbf{Terminus-2} \citep{merrill2026terminalbench} and
\textbf{Codex} \citep{openai2025codex}, two open-source agent harnesses that
adopt the two leaf modes from Section~\ref{sec:construction}:
Terminus-2 uses \hbsmall{} because it has a reliable seed skeleton, and Codex
uses \hblarge{} to infer the hierarchy at repository scale.

Each harness contributes 30 behavior-driven modification requests. They are
grouped by type into \textbf{Query (Q)}, which modifies existing behavior
without revealing target locations; \textbf{Cross-file (CF)}, which adds an
end-to-end capability spanning files or modules; and \textbf{Search-Hostile
(SH)}, which places relevant implementations in locations that are hard to
recover by keyword search. Requests are also labeled \textbf{Easy},
\textbf{Medium}, or \textbf{Hard} by localization difficulty.

\paragraph{\textbf{Evaluation Metrics.}}

We evaluate plan quality, localization accuracy, and planning cost.

\paragraph{Plan quality.}
Three independent judges score each plan on three dimensions:
\emph{Localization} (whether the edit sites are accurate),
\emph{Scope Control} (whether the plan stays focused, with less scope bloat),
and \emph{Reasoning} (whether the rationale and supporting evidence are
adequate). The weighted score is
\begin{equation}
S=
0.5S_{\mathrm{Loc}}
+0.25S_{\mathrm{Scope}}
+0.25S_{\mathrm{Reason}},
\end{equation}
on a 0--100 scale, with Localization receiving the largest weight.
Handbook-Assisted and Baseline are declared the winner when their scores differ
by at least $\delta=3$ points; otherwise the comparison is a tie. The
\textbf{win rate} for any reported slice is the number of wins divided by the
number of valid judge--request comparisons in that slice.

\paragraph{Localization accuracy.}
We compare each predicted edit plan with independent reference plans from
Opus~4.8 and GPT-5.5, reporting Recall, Precision, F1, and Wrong at
file and symbol granularity. Wrong is the share of valid requests with zero
overlap against the reference; lower is better.

\paragraph{Planning cost.}
We report the average number of planner tokens per request, including the
handbook and source context consumed during localization and plan construction.

\subsection{Experiment Results}
\label{sec:exp-results}

The results show three consistent patterns. Handbook guidance improves plan
quality without increasing planner token use. It also helps a weaker planner
produce file- and symbol-level predictions that more closely align with
reference plans from stronger models. These gains persist across request types
and localization difficulty levels. The following subsections examine each
finding in turn.

\subsubsection{Better Plans at Lower Token Cost}
\label{sec:exp-modification}

Figure~\ref{fig:harness-modification-results} summarizes plan-quality win rates
and average planner token use. The Handbook-Assisted arm achieves a higher overall
win rate on both harnesses: 38.3\% versus 28.3\% on Codex and 45.6\% versus
26.7\% on Terminus-2. The improvement is consistent across all three judges: the
gap is 10.0 percentage points for every judge on Codex and ranges from 13.3 to
26.7 points on Terminus-2. Thus, the direction of the result does not depend on
any single judge.

Figure~\ref{fig:behavior-localization-results} decomposes these judgments by
dimension. Averaged across the three judges, the Handbook-Assisted arm has
higher win rates for Localization, Scope Control, and Reasoning on both
harnesses. The respective gains are 12.2, 6.7, and 4.5 points on Terminus-2,
and 2.2, 1.1, and 3.3 points on Codex.

The quality gains are accompanied by lower planner token use. Average use falls
from 0.102M to 0.089M tokens per Codex request (12.7\%) and from 0.058M to
0.053M tokens per Terminus-2 request (8.6\%). The simultaneous increase in win
rate and decrease in token use shows that the improvement is not explained by a
larger planning-token budget.

\subsubsection{A Weaker Planner Matches Stronger Models}

To assess whether handbook guidance helps a weaker planner approach stronger
models in implementation-site localization, we compare its predicted edit sites
with independent reference plans from Opus~4.8 and GPT-5.5
(Table~\ref{tab:answer-key-localization}).

\begin{table}[t]
\centering
\small
\caption{Reference-plan localization metrics (\%). Predicted implementation
sites are compared against independent reference plans generated by Opus 4.8
and GPT-5.5 at file and symbol granularity. Metrics are macro-averaged across
valid requests; missing outputs and planner errors are excluded. \emph{Wrong}
indicates zero overlap with the reference plan at the reported granularity, so
lower values are better.}
\label{tab:answer-key-localization}
\setlength{\tabcolsep}{3pt}
\begin{tabular}{@{}lllccc ccc@{}}
\toprule
& & & \multicolumn{3}{c}{Opus 4.8} & \multicolumn{3}{c}{GPT-5.5} \\
\cmidrule(lr){4-6}\cmidrule(lr){7-9}
Harness & Level & Metric & Baseline & Handbook & Gap & Baseline & Handbook & Gap \\
\midrule
\multirow{8}{*}{Codex}
& \multirow{4}{*}{File} & Recall & 55.4 & 69.7 & +14.3 & 49.4 & 49.8 & +0.4 \\
& & Precision & 46.1 & 60.2 & +14.1 & 53.0 & 62.2 & +9.2 \\
& & F1 & 46.6 & 61.8 & +15.2 & 47.3 & 52.3 & +5.0 \\
& & Wrong $\downarrow$ & 37.0 & 14.8 & -22.2 & 21.4 & 21.4 & 0.0 \\
\cmidrule(l){2-9}
& \multirow{4}{*}{Symbol} & Recall & 47.1 & 65.4 & +18.3 & 46.4 & 49.1 & +2.7 \\
& & Precision & 38.0 & 55.3 & +17.3 & 48.4 & 60.4 & +12.0 \\
& & F1 & 38.3 & 57.1 & +18.8 & 43.8 & 51.2 & +7.4 \\
& & Wrong $\downarrow$ & 44.4 & 18.5 & -25.9 & 28.6 & 21.4 & -7.2 \\
\midrule
\multirow{8}{*}{Terminus-2}
& \multirow{4}{*}{File} & Recall & 74.7 & 83.9 & +9.2 & 76.1 & 87.5 & +11.4 \\
& & Precision & 74.1 & 86.2 & +12.1 & 78.3 & 93.3 & +15.0 \\
& & F1 & 74.1 & 84.7 & +10.6 & 76.5 & 89.3 & +12.8 \\
& & Wrong $\downarrow$ & 24.1 & 13.8 & -10.3 & 20.0 & 6.7 & -13.3 \\
\cmidrule(l){2-9}
& \multirow{4}{*}{Symbol} & Recall & 64.7 & 76.3 & +11.6 & 73.0 & 87.5 & +14.5 \\
& & Precision & 65.1 & 78.4 & +13.3 & 73.9 & 93.3 & +19.4 \\
& & F1 & 64.8 & 77.1 & +12.3 & 73.0 & 89.3 & +16.3 \\
& & Wrong $\downarrow$ & 24.1 & 13.8 & -10.3 & 20.0 & 6.7 & -13.3 \\
\bottomrule
\end{tabular}
\end{table}
Across both harnesses, both reference models, and both granularities, all 24
Recall, Precision, and F1 comparisons are higher for the Handbook-Assisted arm,
with F1 gains ranging from 5.0 to 18.8 points. Recall and precision increase
together, so the improvement is not explained by returning more candidate sites
at the expense of focus.

The absolute agreement is highest on Terminus-2. With handbook guidance, the
weaker planner reaches file-level F1 scores of 84.7\% and 89.3\%, and
symbol-level scores of 77.1\% and 89.3\%, against the Opus~4.8 and GPT-5.5
references, respectively. Precision reaches 93.3\% against GPT-5.5 at both
granularities. The consistent direction across two independent references shows
that the improvement is not specific to one reference model.

Handbook guidance also reduces complete localization failures. Wrong, the share
of requests with zero overlap against the reference, never increases and falls
by as much as 25.9 points. Thus, the gains include fewer complete misses, not
only better overlap among already plausible plans.

\subsubsection{Consistent Gains Across Requests and Difficulty}

Figure~\ref{fig:scenario-generalization-results} stratifies the win-rate results
by request type and localization difficulty.

All six harness-by-type comparisons favor the Handbook-Assisted arm, with gains
ranging from 16.3 to 33.3 percentage points. Codex improves most on Query
requests (26.7 points), whereas Terminus-2 improves most on Search-Hostile
requests (33.3 points). Cross-file requests improve by 16.3 and 20.0 points,
respectively. The gains therefore span existing-behavior changes, cross-file
capabilities, and requests that are difficult to localize through keyword
search.

All six harness-by-difficulty comparisons are also positive, ranging from 3.7
to 33.3 points. Codex improves most on Easy requests, whereas Terminus-2
improves most on Medium requests. Because the gains are not monotonic in labeled
difficulty, the difficulty label alone does not explain their variation.

\section{Conclusion}
\label{sec:conclusion}

This work identifies behavior localization as a central challenge in evolving
production-scale agent harnesses and introduces Harness Handbook, a
behavior-centric representation that links behavioral descriptions to their
distributed source implementations. Harness Handbook reorganizes repository
knowledge around runtime behavior, constructs this representation directly from
source, and supports modification through behavior-guided progressive
disclosure and automatic resynchronization.

Our experiments demonstrate the value of this representation along three
dimensions. First, handbook guidance improves plan quality while lowering
planning cost: on Codex and Terminus-2, overall win rates rise by 10.0 and 18.9
percentage points, while planner token use falls by 12.7\% and 8.6\%,
respectively. Second, with the handbook a weaker planner matches the
implementation-site localization of substantially stronger models, improving all
24 file- and symbol-level Recall, Precision, and F1 comparisons against two
independent reference plans. Third, these gains persist across request types and
localization difficulty levels. Together, these results indicate that making the
relationship between behavior and implementation explicit can improve both
modification planning and localization efficiency without requiring a larger
planning-token budget.

The utility of Harness Handbook extends beyond modification planning. As a
behavior-centric repository representation that stays synchronized with the code,
it can also support broader uses such as behavior auditing and regression-impact
analysis. Our next step is to apply it to harness self-evolving: using the
handbook as a shared behavioral memory, an agent can autonomously close the loop
of localization, planning, execution, and resynchronization as the repository
evolves, moving the harness toward self-improvement.

\clearpage
\nocite{*}
\bibliographystyle{unsrtnat}
\bibliography{references}

\clearpage
\appendix

\etocsettocdepth.toc{subsubsection}
\begingroup
\hypersetup{linkcolor=black}
\etocsetnexttocdepth{subsubsection}
\renewcommand{\contentsname}{Appendix Contents}
\etocsettocstyle{\section*{\contentsname}\vspace{10pt}}{}
\linespread{1.5}\selectfont
\tableofcontents
\endgroup
\clearpage

\section{Harness Handbook Construction Details}
\label{app:construction-details}

This appendix expands the three construction phases in
Section~\ref{sec:construction}. The selected leaf mode $g$ and construction
configuration $\Theta$ stay fixed through construction and later
resynchronization. Both modes work from the same static facts but organize them
differently: \hbsmall{} starts from a trusted seed skeleton $\mathcal{S}_0$ and
places functions within it, whereas \hblarge{} describes files first and infers
the skeleton. Both ultimately produce the same reader-facing handbook---an
L1--L3 document tree, a cross-stage state-register view, and source locators for
L3 entries. Algorithm~\ref{alg:construction} summarizes the shared workflow and
its mode-specific branches.

\begin{algorithm}[H]
\caption{Harness Handbook Construction}
\label{alg:construction}
\begin{minipage}{0.96\linewidth}
\footnotesize
\renewcommand{\baselinestretch}{0.96}\selectfont
\begin{algorithmic}[1]
\Require repository $\mathcal{R}$; leaf mode
$g\in\{\mathrm{function},\mathrm{file}\}$; construction configuration $\Theta$;
seed skeleton $\mathcal{S}_0$ if $g=\mathrm{function}$ (otherwise $\emptyset$)
\Ensure handbook $\mathcal{H}$ containing rendered view $\mathcal{V}$, document
tree $\mathcal{D}$, register view $\mathcal{Z}$, fixed leaf mode $g$, and
synchronization state $\mathcal{K}_g$
\Statex \textit{Phase I: shared static facts.}
\State $\mathcal{G}\gets\Call{BuildGraph}{\mathcal{R}}$
\Comment{deterministic; no LLM}
\Statex
\Statex \textit{Phase II: mode-specific behavioral organization.}
\If{$g=\mathrm{function}$}
    \State $(\mathcal{S},\mathcal{U}_g)\gets
    \Call{OrganizeFunctions}{\mathcal{R},\mathcal{G},\mathcal{S}_0,\Theta}$
    \Comment{budgeted stage and membership review}
    \State $\mathcal{U}_g\gets
    \Call{FinalizeFunctions}{\mathcal{R},\mathcal{G},\mathcal{S},\mathcal{U}_g}$
    \Comment{regions, ordering, and unmapped functions}
\Else \Comment{$g=\mathrm{file}$}
    \State $(\mathcal{S},\mathcal{U}_g)\gets
    \Call{OrganizeFiles}{\mathcal{R},\mathcal{G},\Theta}$
    \Comment{cards, stage inference, and assignments}
    \If{$\Call{RefinementEnabled}{\Theta}$}
        \State $(\mathcal{S},\mathcal{U}_g)\gets
        \Call{RefineFileOrganization}{\mathcal{R},\mathcal{G},\mathcal{S},\mathcal{U}_g,\Theta}$
    \EndIf
    \State $\mathcal{U}_g\gets
    \Call{FinalizeFiles}{\mathcal{G},\mathcal{S},\mathcal{U}_g}$
    \Comment{within-stage order and coverage}
\EndIf
\Statex
\Statex \textit{Phase III: synthesis, grounding, and packaging.}
\State $(\mathcal{D},\mathcal{Z},\mathcal{B})\gets
\Call{Synthesize}{g,\mathcal{R},\mathcal{G},\mathcal{S},\mathcal{U}_g,\Theta}$
\State $\mathcal{Z}\gets\Call{ValidateRegisterRefs}{\mathcal{Z},\mathcal{S}}$
\State $\mathcal{D}\gets
\Call{ValidateAndFreezeLocators}{\mathcal{D},\mathcal{R},g}$
\State $\mathcal{K}_g\gets
\Call{Archive}{g,\mathcal{G},\mathcal{S}_0,\mathcal{S},\mathcal{U}_g,\mathcal{B},\Theta}$
\Comment{$\mathcal{S}_0$ is stored only in function mode}
\State $\mathcal{H}\gets
\Call{RenderAndPackage}{\mathcal{D},\mathcal{Z},g,\mathcal{K}_g}$
\State \Return $\mathcal{H}$
\end{algorithmic}
\end{minipage}
\end{algorithm}

\subsection{Phase I: Shared Static Fact Extraction}

Phase~I is a single deterministic analysis shared by both modes.
Language-specific adapters parse $\mathcal{R}$ and build the program graph
$\mathcal{G}$. An internal-function node records a function's qualified name,
source file, signature, line range, enclosing class, and observed state access;
a boundary node represents a named external target, such as a library function.

The graph keeps a call edge only when the target resolves to an internal
function or a named boundary; unresolved calls are written to an audit log
rather than assigned a guessed target. Because this phase uses no LLM, every
later organization step builds on the same source-grounded facts.

\subsection{Phase II: Behavioral Organization}

Phase~II organizes the facts in $\mathcal{G}$ around execution behavior. Its
output is the current stage skeleton $\mathcal{S}$ together with the
mode-specific organization state $\mathcal{U}_g$. The two leaf modes differ in
both their starting point and their unit of organization.

\subsubsection{Function-as-Leaf}

\paragraph{Initial assignment.}
The \hbsmall{} branch ($g=\mathrm{function}$) starts from the seed skeleton
$\mathcal{S}_0$, whose stage and state-register declarations initialize
$\mathcal{S}$. On the first pass, it considers every analyzable
internal function---that is, every non-synthetic function with usable source
locations---and attempts to assign it to one or more stages. The decision uses
the function's source in $\mathcal{R}$, its callers and callees in $\mathcal{G}$,
the current neighboring assignments, and the stage descriptions. A function is
normally placed as a whole. If different parts serve different behavioral roles,
it can instead be divided into contiguous source regions and each region placed
separately.

\paragraph{Review and convergence.}
A single pass may expose weaknesses in either the assignments or the seed
structure, so later rounds review both. Each round first checks whether stages
should be added, removed, merged, or split. A structural change is accepted
only if the resulting hierarchy is valid and every state-register declaration
still refers to an existing stage. Functions affected by an accepted change are
then classified again against the revised structure.

After reclassification, a global audit checks whether whole-function placements
remain consistent with the surrounding stages, and a separate pass repairs the
boundaries of retained regions. Proposed changes must pass model-based
review before they are applied. The rounds stop when the stage skeleton and all
function or region assignments match the
previous round and no function is waiting to be reconsidered. If this state is
not reached, the process stops at the organization budget stored in $\Theta$ and
keeps the remaining gaps explicit.

\paragraph{Validation and finalization.}
Every accepted assignment must name a valid stage, use a legal source range, and
avoid conflicting actions. Region boundaries are aligned to parser-derived
statement spans when possible, and the enclosed source is hashed for later
comparison. A boundary that cannot be aligned is marked \emph{needs-review};
Phase~III then decides whether its locator can remain active or must be frozen.

Deterministic cleanup removes a redundant whole-function entry when retained
regions already cover that function in the same stage, without disturbing its
assignments to other stages. The pipeline then rebuilds cross-stage references,
records all functions that remain unmapped, and orders entries within each stage
using their purposes and stage-internal calls. If the proposed order is incomplete
or contains duplicates, file and line order provides a stable fallback. Together,
the finalized memberships, region anchors and hashes, boundary status, within-stage
order, and record of unmapped functions constitute $\mathcal{U}_{\mathrm{function}}$.

\subsubsection{File-as-Leaf}

The \hblarge{} branch ($g=\mathrm{file}$) uses no seed skeleton
($\mathcal{S}_0=\emptyset$). It first describes each file, then infers
$\mathcal{S}$ from those descriptions, and finally organizes the files within
each stage.

\paragraph{Step 1: Build a card for each file.}
The pipeline attempts to create one file card for every scanned file,
including files from which static analysis extracts no function. A successfully
generated card describes the file's purpose, repository role, and place in the
execution lifecycle. When deep card generation is enabled, it also includes a
static inventory of function identities, signatures, source ranges, and resolved
calls. The model may explain this inventory but cannot change it.

Card generation uses a fixed fallback sequence. A file omitted from a batch is
retried individually. If a deep card still fails, the source is retried in
function-range chunks. If no prose can be produced, a fallback card is emitted;
in deep mode, it retains the graph-derived inventory. The file stays available to
later steps but is marked undescribed in the coverage record $\mathcal{Y}$, so no
scanned file is silently dropped.

\paragraph{Step 2: Infer stages and assign files.}
The pipeline groups cards by directory and combines their descriptions with
graph-derived entry points and resolved calls. From this context, it drafts an
ordered stage skeleton and attempts to assign each scanned file to
exactly one primary stage, using its card description when available. That stage
is the file's home in the handbook. A genuinely
cross-cutting file, such as a logging, protocol, type, or configuration utility,
may also name one or two secondary stages. Missing assignments and assignments to
unknown stages remain visible in $\mathcal{Y}$ rather than being forced into the
nearest stage.

The organization variant stored in $\Theta$ determines whether the initial
result is refined.
\texttt{oneshot}, the default, ends after one stage draft and one file-assignment
pass. \texttt{doctor} starts from that draft and applies a reviewed refinement
loop. \texttt{agent} uses an agent-generated draft followed by the same loop; if
the agent cannot produce a valid draft, it falls back to the one-shot draft. A
refinement round may add, remove, merge, or split stages, after which only
affected or unassigned files are reconsidered. The loop stops when all files are
assigned and the skeleton is stable, when neither coverage nor stage balance
improves for two consecutive rounds, or when the organization budget is reached.
Any files that remain unassigned stay visible in $\mathcal{Y}$.

\paragraph{Step 3: Organize files within each stage.}
For each non-empty stage, the file-call graph first provides a
caller-before-callee order. The file cards are then used to refine that order into
thematic groups and a readable sequence. A deterministic check removes unknown or
duplicate paths and verifies that every assigned file appears exactly once. An
omitted file is appended to a fallback group, and a failed organization call falls
back to the deterministic flat order. This step can reorder files within a stage
but cannot change the stage skeleton. Together, the file cards, stage
assignments, within-stage grouping and order, and coverage record $\mathcal{Y}$
constitute $\mathcal{U}_{\mathrm{file}}$.

\subsection{Phase III: Hierarchical Synthesis and Packaging}

Phase~III turns $(\mathcal{S},\mathcal{U}_g)$ into the document tree
$\mathcal{D}$ and state-register view $\mathcal{Z}$, records reusable generation
outputs in the cache $\mathcal{B}$, and validates both against the repository
before packaging.

\paragraph{Synthesizing the document tree.}
In both modes, $\mathcal{D}$ contains an L1 system overview, an L2 component
overview for each included stage, and source-backed L3 entries. The modes differ
in the direction of synthesis:

\begin{itemize}[leftmargin=1.4em,itemsep=2pt,topsep=2pt]
\item \textbf{\hbsmall{} builds top-down.} The known stage skeleton provides
the outline. Generation proceeds from the L1 overview to each L2 component
overview and then to the stage's L3 entries. Each L3 entry represents one function within
that stage; multiple retained regions of the same function are merged into that
entry.
\item \textbf{\hblarge{} builds bottom-up.} Assigned file cards become the L3
entries. A stage is summarized when it owns assigned files or has child stages;
empty leaf placeholders are skipped. The remaining stages are processed from the
leaves upward: child summaries and one-line descriptions of a stage's own files
produce its L2 component overview, and the top-level summaries produce L1.
\end{itemize}

In \hbsmall{} mode, each L1, L2, and L3 node uses a bounded
generate--review--revise process. The best draft is retained, and generation
stops when it passes the rubric, its score no longer improves, or the generation
budget is reached. Independent stages can be processed in parallel, but L3
entries within one stage are generated in sequence so that a later entry can
refer to earlier ones.

In \hblarge{} mode, file cards are rendered directly. Each stage and system
summary is generated once, with a deterministic fallback if generation fails,
and each parent summary waits for its child summaries.

The resulting descriptions and summaries are retained in $\mathcal{B}$ so that
unchanged content can be reused during later generation or resynchronization.

\paragraph{Building the state-register view.}
In \hbsmall{} mode, the register identities and stage relationships in
$\mathcal{Z}$ come from the declarations in the current stage skeleton;
generation turns these declarations into reader-facing descriptions without
inventing new registers.
In \hblarge{} mode, the pipeline instead proposes cross-stage registers from the
top-level stage summaries and the purposes of data-model files. It repeats this
search until two consecutive rounds add nothing or the register budget is
reached. References to stages absent from the current structure are removed from
the candidate mapping.

\paragraph{Grounding and failure handling.}
The static facts in $\mathcal{R}$ and $\mathcal{G}$ remain authoritative
throughout synthesis. The model may add behavioral explanations, but it cannot
alter file paths, function identities, signatures, source ranges, or resolved
calls. Functions and files that cannot be placed remain explicitly recorded
rather than being assigned by guesswork.

Before rendering, the pipeline resolves every L3 locator in $\mathcal{D}$
against $\mathcal{R}$ and compares the located source with the stored evidence.
An entry that cannot be revalidated is marked frozen and excluded from
localization until it is refreshed, keeping the repository authoritative even
when prose is reused from the cache.

\paragraph{Rendering and saved synchronization state.}
After validation, the pipeline renders $\mathcal{D}$ and $\mathcal{Z}$ as the
reader-facing view $\mathcal{V}$. It then packages this view with $\mathcal{D}$,
$\mathcal{Z}$, the fixed $g$, and the machine-readable resynchronization state
$\mathcal{K}_g$ as the handbook $\mathcal{H}$. Across both modes, $\mathcal{K}_g$
retains the fixed leaf mode, program graph, current stage skeleton, mode-specific
organization state, construction configuration, and generation cache. The
stored configuration $\Theta$ records the selected organization or refinement
variant, card-detail level, and the budgets for organization, generation, and
register extraction where applicable. In function mode,
$\mathcal{K}_g$ also retains the original seed skeleton $\mathcal{S}_0$; no seed
is stored in file mode. Saving $\mathcal{H}$ in this form allows a later code
change to refresh affected content without reconstructing unchanged parts of
the handbook.

\section{Modification and Resynchronization Details}
\label{app:modification-details}

This appendix expands the four-step modification workflow in
Section~\ref{sec:modification} using the package contract defined in
Appendix~\ref{app:construction-details}. It follows a request $q$ through BGPD
localization, edit planning, execution, and automatic resynchronization.
Throughout the workflow, the handbook $\mathcal{H}$ is paired with its matching
repository $\mathcal{R}$, and the stored leaf mode $g$ remains fixed.

\subsection{Behavior-Guided Progressive Disclosure}

BGPD guides a read-only agent through the handbook from coarse to fine. A
locator enters the evidence set only after it has been resolved against the
current repository.

\paragraph{Stage selection.}
Navigation begins at L1 and L2 of the document tree $\mathcal{D}$ stored in
$\mathcal{H}$. The agent reads the system overview and stage index to select the
execution stages whose behavior matches $q$. It then follows the state-register
view $\mathcal{Z}$ to add stages coupled through shared state. This step recovers
behavior that is structurally distant yet mutually dependent, such as a value
written in one stage and consumed several stages later.

\paragraph{Entry selection.}
Within the selected stages, the agent opens the corresponding stage pages and
chooses the most relevant L3 entries. Each entry exposes a summary and a source
locator. Its body is disclosed only when needed, limiting unnecessary context.

\paragraph{Call-relation expansion.}
The candidate set is expanded along call relations in the program graph
$\mathcal{G}$ retained by the synchronization state $\mathcal{K}_g$. In
\hbsmall{} mode, expansion follows the function-call graph; in \hblarge{} mode,
it follows the induced file-call graph. Named external boundary nodes may supply
context but are never returned as edit sites.

\paragraph{Source verification.}
All preceding steps operate on $\mathcal{H}$. The agent then opens
$\mathcal{R}$, resolves each candidate locator, and keeps only the sites that
remain relevant to $q$. The result is the verified evidence
$\widehat{\mathcal{E}}_q$. Each record carries a file path, an optional function
or region anchor, and a current source excerpt. The handbook guides the search,
while the repository remains the authority for the edit plan.

\subsection{Edit Planning}

The planner converts $\widehat{\mathcal{E}}_q$ into an edit plan $\mathcal{P}$
whose blocks name a target file, an optional anchor, the supporting source
excerpt, and the intended change. In our evaluation implementation, each change
is expressed as a verbatim old/new pair. For every block, the planner also emits
an action declaration. These declarations are collected as
$\Gamma=(\Gamma_{\mathrm{modify}},\Gamma_{\mathrm{add}},
\Gamma_{\mathrm{remove}})$, with a rename encoded as one removal and one
addition.

\subsection{Execution}

Execution is delegated to a separate agent that applies $\mathcal{P}$ to
$\mathcal{R}$ and produces $\mathcal{R}'$. The factual record of the change is
$\Delta=\mathrm{Diff}(\mathcal{R},\mathcal{R}')$.

In our implementation, the executor can read and write files, perform targeted
replacements, and search within a file, but it cannot list directories or access
a shell. It first applies each verbatim replacement directly and reads the target
source only if the replacement fails. It then verifies the touched files and runs
a target-specific syntax gate when one is configured. The diff $\Delta$ drives
handbook invalidation, whereas $\Gamma$ is used only to check whether execution
followed the plan.

\subsection{Automatic Resynchronization}

Every non-empty $\Delta$ automatically invokes $\mathrm{Resync}_g$, which updates
only the affected parts of the handbook whenever possible. If
$\Delta=\emptyset$, resynchronization is skipped and $\mathcal{H}'=\mathcal{H}$.

\paragraph{Version alignment.}
The procedure reparses $\mathcal{R}'$ and builds the updated program graph
$\mathcal{G}'$. It compares $\mathcal{G}'$ with the stored graph $\mathcal{G}$,
using $\Delta$ to identify added, removed, and modified units.

In \hbsmall{} mode, functions are matched using body fingerprints that ignore
line numbers. A function that only moves is therefore treated as unchanged, and
its locator is shifted by the corresponding line offset. A rename is detected by
matching the body below the signature line. In \hblarge{} mode, file-set
differences identify added and removed files, while content hashes identify
changed files. A path rename therefore appears as one removal and one addition.

Before the observed actions are compared with $\Gamma$, a detected function
rename is normalized to one removal and one addition. This audit does not alter
the $\Delta$-driven invalidation scope.

\paragraph{Scoped update.}
Resynchronization next checks whether the current stage skeleton $\mathcal{S}$
can accommodate the change. If it remains valid, $\mathcal{S}'=\mathcal{S}$ and
the affected part of $\mathcal{U}_g$ is updated to $\mathcal{U}'_g$. In
\hbsmall{} mode, this update removes deleted memberships, carries forward
matched functions, and reclassifies new or changed functions. It also refreshes
region anchors, source hashes, within-stage order, and unmapped status. In
\hblarge{} mode, it removes obsolete cards and assignments, refreshes cards for
new or changed files, assigns new files, reorganizes affected stages, and updates
the coverage record $\mathcal{Y}$ to $\mathcal{Y}'$.

The pipeline then regenerates the affected L3 entries, their enclosing L2/L1
ancestors, and dependent state registers. This produces the updated document
tree $\mathcal{D}'$ and state-register view $\mathcal{Z}'$. Unaffected generation
outputs are reused from $\mathcal{B}$. The updated cache $\mathcal{B}'$ retains
these reusable entries and adds or replaces the refreshed outputs.

If $\mathcal{S}$ can no longer accommodate the change,
Algorithm~\ref{alg:construction} is rerun on $\mathcal{R}'$ with the same leaf
mode $g$ and stored configuration $\Theta$. The \hbsmall{} branch also reuses the
stored seed skeleton $\mathcal{S}_0$. The \hblarge{} branch instead infers a new
stage skeleton from the updated repository. During the full rebuild, Phase~I
produces $\mathcal{G}'$, Phase~II produces
$(\mathcal{S}',\mathcal{U}'_g)$, and Phase~III synthesis produces
$(\mathcal{D}',\mathcal{Z}',\mathcal{B}')$. These components follow the shared
validation and packaging contract described below.

\paragraph{Conservative handling.}
Source that cannot be parsed or classified is not guessed. An existing L3 entry
whose locator or source hash cannot be revalidated is marked frozen and excluded
from localization. In \hbsmall{} mode, new or changed functions that cannot be
mapped remain explicitly recorded as unmapped in
$\mathcal{U}'_{\mathrm{function}}$. If a changed function previously had an
active L3 entry, that entry is frozen until classification succeeds. In
\hblarge{} mode, files without a valid card or assignment remain in the coverage
record $\mathcal{Y}'$ within $\mathcal{U}'_{\mathrm{file}}$.

Parsing, hashing, graph construction, locator roll-forward, and coverage checks
are deterministic. Model calls are limited to classification, file assignment,
within-stage organization, and description revision. These categories also
cover a full rebuild. Stage and function decisions count as classification;
file-to-stage decisions count as file assignment; and file grouping counts as
within-stage organization. Regenerating cards, document nodes, or registers
counts as description revision.

\paragraph{Validation and packaging.}
Both update paths then follow the packaging contract of
Appendix~\ref{app:construction-details}: the pipeline validates the locators in
$\mathcal{D}'$ against $\mathcal{R}'$ and the stage references in $\mathcal{Z}'$
against the updated skeleton, archives $\mathcal{G}'$ and
$(\mathcal{S}',\mathcal{U}'_g)$ with $\mathcal{B}'$, the fixed $g$, and $\Theta$
(plus $\mathcal{S}_0$ in function mode) as $\mathcal{K}'_g$, and renders and
packages the validated $(\mathcal{D}',\mathcal{Z}')$ with $\mathcal{K}'_g$ as
$\mathcal{H}'$. The pair $(\mathcal{R}',\mathcal{H}')$ becomes the starting state
for the next request.

\section{Experimental Details}
\label{app:experimental-protocol}

This appendix details the setup summarized in Section~\ref{sec:exp-setup}, where
every factor other than handbook access is held fixed across the two arms. It
covers the two evaluated harnesses and their Phase~I static facts, the
modification requests ($30$ per harness, balanced across request type and
localization difficulty), and the two evaluation arms with the \texttt{SKILL.md}
manifest that exposes the handbook to the planner. The evaluation metrics are
defined in Section~\ref{sec:exp-setup}.

\subsection{Harnesses and Static Facts}

We evaluate on two production-style agent harnesses that differ sharply in scale
and language.

\begin{itemize}[leftmargin=1.4em,itemsep=2pt,topsep=2pt,parsep=0pt]
\item \textbf{Terminus-2} is a Python terminal agent from the Harbor framework. It
drives a real terminal through a tmux session in an observe--decide--act loop:
it reads the terminal state, queries an LLM for the next action, sends commands,
and reads back the results, repeating until the task finishes or a limit is
reached. It is compact (6 source files) but behaviorally rich, with a multi-stage
iteration loop, context management, and cross-iteration state.
\item \textbf{Codex} is the Rust monorepo behind the Codex coding agent. It is
large and multi-surface, spanning a CLI, a TUI, an app-server, configuration, and
sandboxing across many crates, with thousands of files and deep call graphs.
\end{itemize}

Following the selection rule in Section~\ref{sec:construction}, Terminus-2 uses the
\hbsmall{} mode, since a reliable seed skeleton is available and function-level
entries stay within budget, whereas Codex uses the \hblarge{} mode to infer the
hierarchy at repository scale. Table~\ref{tab:handbook-stats} reports the Phase~I
static facts extracted from each repository together with the resulting handbook
structure.

\begin{table}[h]
\centering
\small
\caption{Phase~I static facts and resulting handbook structure for the two
evaluated harnesses. The upper block reports the program graph $\mathcal{G}$
produced by static analysis (Appendix~\ref{app:construction-details}); the lower
block reports the built handbook. For Terminus-2, an L3 entry covers a whole
function or one or more contiguous regions of a function; for Codex, an L3 entry
corresponds to one file.}
\label{tab:handbook-stats}
\setlength{\tabcolsep}{10pt}
\begin{tabular}{@{}l l r r@{}}
\toprule
& & \textbf{Terminus-2} & \textbf{Codex} \\
\midrule
\multirow{2}{*}{\emph{Setup}}
  & Language  & Python     & Rust \\
  & Leaf mode & \hbsmall{} & \hblarge{} \\
\midrule
\multirow{4}{*}{\parbox[c]{2.3cm}{\emph{Program graph\\(Phase~I)}}}
  & Source files            & 6   & 2{,}267 \\
  & Internal-function nodes & 103 & 34{,}363 \\
  & Boundary nodes          & 41  & 4{,}016 \\
  & Resolved call edges     & 257 & 159{,}960 \\
\midrule
\multirow{3}{*}{\parbox[c]{2.3cm}{\emph{Built\\handbook}}}
  & Stages (L2)     & 20  & 140 \\
  & L3 entries      & 106 & 2{,}267 \\
  & State registers & 10  & 62 \\
\bottomrule
\end{tabular}
\end{table}

The two repositories differ by roughly two orders of magnitude in files,
functions, and call edges, which is precisely why they instantiate the two leaf
modes.

\subsection{Modification Requests}

Each harness contributes 30 behavior-driven modification requests, evenly split
into three types. Each request states an intended behavioral change without
revealing target locations. The two suites are built from the same behavioral
intents but phrased against each harness's own code, so the same goal lands on
different implementation sites in Terminus-2 and Codex.

\begin{itemize}[leftmargin=1.4em,itemsep=2pt,topsep=2pt,parsep=0pt]
\item \textbf{Query (Q)} modifies existing behavior, such as a trigger, a timing
rule, or a control-flow decision.
\item \textbf{Cross-file (CF)} adds an end-to-end capability that spans files,
schemas, pipeline logic, or interfaces.
\item \textbf{Search-Hostile (SH)} places relevant sites in mirrored
implementations, fallback branches, or cold paths that keyword search alone
struggles to recover.
\end{itemize}

Table~\ref{tab:request-examples} gives a representative request of each type,
chosen separately for each harness.

\begin{table}[!ht]
\centering
\footnotesize
\caption{Representative modification requests, chosen separately for each harness,
with the localization challenge each type poses.}
\label{tab:request-examples}
\setlength{\tabcolsep}{5pt}
\renewcommand{\arraystretch}{1.08}
\begin{tabularx}{\textwidth}{@{}l Y Y Y@{}}
\toprule
& \textbf{Terminus-2} & \textbf{Codex} & \textbf{Localization challenge} \\
\midrule
\textbf{Q} & Require marking \texttt{task\_complete} three times, with an ``are you
sure?'' re-prompt, before grading. & Raise \texttt{exec\_command}'s wait ceiling to
300\,000\,ms, keeping schema bound and docs in agreement. & An existing behavior
must be found and altered; the difficulty is telling the target apart from similar
triggers or rules. \\
\addlinespace[2pt]
\textbf{CF} & Add a \texttt{background} flag that sends keystrokes without capturing
output, across the terminal layer and both formats. & Add a per-command
\texttt{env} override, threaded through the spawn path across both shell tools. &
The capability spans coordinated sites (schema, pipeline, interface, docs) that
must all be located together. \\
\addlinespace[2pt]
\textbf{SH} & Mask secrets before any model snapshot and before writing the session
recording. & Mask secrets in every capture path and before writing the on-disk
rollout. & Relevant sites hide in mirrored, fallback, or cold paths, so keyword
search alone misses some of them. \\
\bottomrule
\end{tabularx}
\end{table}

Requests are additionally labeled Easy, Medium, or Hard by localization
difficulty, reflecting whether the change touches one primary behavior, requires
coordination across stages and files, or requires discovering dependencies along
indirect execution paths. Table~\ref{tab:difficulty-examples} gives a
representative request at each level for both harnesses.

\begin{table}[!ht]
\centering
\footnotesize
\caption{Representative modification requests at each localization-difficulty
level, chosen separately for each harness, with the localization challenge each
level poses.}
\label{tab:difficulty-examples}
\setlength{\tabcolsep}{5pt}
\renewcommand{\arraystretch}{1.08}
\begin{tabularx}{\textwidth}{@{}l Y Y Y@{}}
\toprule
& \textbf{Terminus-2} & \textbf{Codex} & \textbf{Localization challenge} \\
\midrule
\textbf{Easy} & Strip ANSI escape codes from all captured terminal output. & Add an
informational \texttt{comment} field on both shell tools. & One well-defined site:
the change maps to a single behavior that a keyword or one file read can pinpoint. \\
\addlinespace[2pt]
\textbf{Medium} & Add a per-command \texttt{cwd}, threaded through execution, the
terminal layer, and both formats. & Add an \texttt{expect} output check on both
shell tools, annotating the observation on failure. & Coordinated edits: one field
or check must be threaded through parsing, execution, and documentation, so several
related sites must be found together. \\
\addlinespace[2pt]
\textbf{Hard} & Keep run counters and step history consistent across a mid-run
compaction. & Reconcile overlapping fields across \texttt{exec\_command},
\texttt{shell\_command}, and \texttt{write\_stdin}. & Indirect coupling: the sites
sit on cold or mirrored paths (post-compaction bookkeeping; parallel tool schemas)
that keyword search rarely surfaces. \\
\bottomrule
\end{tabularx}
\end{table}

\begin{figure}[t!]
\begin{lstlisting}[style=skillbox]
---
name: <harness>-handbook
description: >-
  Structural map of the harness, derived from its code. Use it when planning a
  change to find EVERY code site the change must touch -- it maps the code
  stage-by-stage and lists every state register together with all of its
  read/write locations. Consult it before finalizing a plan so that scattered or
  non-obvious sites are not missed.
---

# Harness Handbook: navigation guide

This is a structural map of the harness. Use it to locate where a requested change
must take effect -- especially sites that are not adjacent to the obvious one.

## Reference files
- references/overview.md    -- whole-system orientation: the lifecycle, the main
                               loop, and how the stages fit together. Read first.
- references/index.md       -- the map: every stage (id, title, what it does, the
                               leaves it covers) and every state register.
- references/registers.md   -- for each state register: its purpose and EVERY place
                               it is written and read, across the whole harness.
- references/stages/<id>.md -- one stage's prose plus a card per leaf (function or
                               file): its <summary> line gives the exact code
                               location (path:start-end), and the card holds the
                               interface, behavior, relations, and source.

## How to use it (during planning, before you write your plan)
1. Read references/overview.md first to understand the whole system.
2. Read references/index.md to identify the stages, leaves, AND state registers
   your change involves. Do NOT prematurely narrow to the one obvious stage: a
   single change often touches sites in several stages.
3. For EVERY state register your change touches, read references/registers.md and
   note EVERY write and read site. This read/write registry surfaces scattered,
   non-adjacent sites a top-down code read would miss.
4. Open the relevant references/stages/<id>.md for detail and coupled assumptions.
5. For each site the handbook names, open the real code with read_file /
   search_file_content and locate the precise lines.
6. Your plan must account for every site the handbook surfaced.

The handbook tells you WHERE things live and how they connect. You still decide
what the change should be, and you verify every location against the real code
before planning it.
\end{lstlisting}
\caption{The \texttt{SKILL.md} manifest that exposes a handbook to the planner in the Handbook-Assisted arm. The same structure is used in both leaf modes; a leaf is a function in \hbsmall{} mode and a file in \hblarge{} mode.}
\label{lst:skill}
\end{figure}

\subsection{Evaluation Arms}

To isolate the effect of the handbook, both arms share the same configuration. The
planner is an agent built on NexAU whose underlying LLM is DeepSeek-V4-Pro and
whose repository tools are read-only, and both arms use the same requests,
repository snapshot, planning contract, and decoding settings. They differ only in
whether the handbook is available and in the corresponding localization
instructions. The assisting handbook is built from the same repository snapshot
used in the evaluation.

\begin{itemize}[leftmargin=1.4em,itemsep=1pt,topsep=1pt,parsep=0pt]
\item \textbf{Baseline.} Without a handbook, the planner locates edit sites by
exploring the repository with read-only tools (file read, in-file search, and
directory listing).
\item \textbf{Handbook-Assisted.} The planner is additionally given the handbook as
a navigable skill and follows the BGPD policy, proposing source locators from the
handbook and then verifying them against the repository.
\end{itemize}

The handbook is exposed to the planner as a skill: a \texttt{SKILL.md} manifest
tells the planner when and how to consult the handbook, and points to the
reference files it navigates. Figure~\ref{lst:skill} shows this manifest. The
same structure is used in both leaf modes; each stage's leaf cards hold functions
in \hbsmall{} mode and files in \hblarge{} mode.

\begingroup
\makeatletter
\setlength{\@fptop}{0pt}
\setlength{\@fpbot}{0pt plus 1fil}
\makeatother
\clearpage
\endgroup
\section{Prompt Templates}
\label{app:prompts}

This appendix reproduces the main prompts used by the pipeline, organized by the
stage in which they are used: those that build the handbook
(Appendix~\ref{sec:d-generation}), those that localize and plan a change under
BGPD (Appendix~\ref{sec:d-bgpd}), and those that score plans with an LLM judge
(Appendix~\ref{sec:d-judge}). The listings are transcribed from the
implementation; per-run fields substituted at runtime are shown as
\texttt{\{\{...\}\}} or \texttt{\{...\}} placeholders, and a few non-ASCII
characters are rendered in ASCII. Static fields such as qualified names,
signatures, source ranges, and call edges are supplied by static analysis and
cannot be overwritten by the model.

\subsection{Handbook Generation}
\label{sec:d-generation}

These prompts drive the construction pipeline of
Appendix~\ref{app:construction-details}: they organize source units into
execution stages and synthesize the L1--L3 document tree. Throughout, the
statically extracted identities stay fixed and the model only supplies behavioral
descriptions.

\subsubsection{Classifying Functions into Stages}

Used in \hbsmall{} Phase~II to assign one function to one or more stages, as a
whole or as contiguous regions. The rules block below is followed at runtime by
the available stage list, the function's metadata, its caller/callee context with
their current stage assignments, and the line-numbered source.

\begin{lstlisting}[style=promptbox]
You are analyzing one Python function from the Terminus 2 agent harness, and
classifying it into one or more stages of a hand-authored data-flow skeleton.

GRANULARITY
- "function": the entire function is a single narrative unit. Use for short,
  cohesive functions (<=30 lines or with a single clear purpose).
- "region": the function contains multiple distinct narrative steps and should be
  split into 2-10 regions. Use for large functions with multiple decision points
  or sequential phases.

REGION RULES (when granularity == "region")
- Each region MUST be contiguous and end at a complete statement boundary (not
  mid-statement).
- Provide first_line and last_line as the EXACT text content of the first and last
  lines of the region.
- Each region has its own stage_id.

STAGE ASSIGNMENT
- Pick stage IDs ONLY from the list provided.
- CROSS-CUTTING utility functions (small utilities called from many places -- token
  counting, output length capping, recording markers) go to crosscut-X* ONLY. Do
  NOT add them to consuming stages.
- A function that uses self.logger.X is NOT cross-cutting just because of that --
  assign by its primary identity.
- For api-surface methods (e.g. name(), version()): set function_assignments=[] and
  granularity="function" (will be recorded as unmapped api_surface).
- For subsystem-internal functions (in tmux_session.py / terminus_*_parser.py /
  asciinema_handler.py): assign to the main-flow stage that drives them.

PURPOSE FIELD (60-150 words, structured across 5 aspects)
1. ACTION: what the function does, concretely (no "handles" / "manages")
2. INPUTS / STATE READ: arguments + self._* attrs that determine behavior
3. OUTPUTS / STATE WRITTEN: return value + self._* attrs mutated
4. WHEN INVOKED: who calls it, under what condition
5. NON-OBVIOUS: retry logic, fallback paths, design choices a reader would miss

CONSISTENCY WITH CONTEXT
Look at the caller/callee context provided -- your classification should be
consistent with their stage assignments. A function that's only called from
stage-4.2's region typically belongs near stage-4.2.

OUTPUT
Return ONLY a JSON object inside a ```json fenced block:

{
  "qualname": "<exact qualname>",
  "purpose": "<60-150 word 5-aspect description>",
  "granularity": "function" | "region",
  "function_assignments": ["stage-X", ...],
  "regions": [
    {"line_range":[a,b], "first_line":"...", "last_line":"...",
     "purpose":"<30-80 word>", "stage_id":"..."}, ...
  ] | null,
  "file": "<file>",
  "line_range": [<start>, <end>]
}

When granularity == "function", regions = null.
\end{lstlisting}

\subsubsection{Reviewing Function Assignments}

Each proposed assignment is reviewed by a critic against the source, the stage
definitions, and caller/callee evidence. The reviewer role block (the ``engineer''
role) is followed by the task context, the review evidence, and the proposal under
review, then the output rules below. Reviewers lean toward APPROVE so the pipeline
keeps making progress.

\begin{lstlisting}[style=promptbox]
You are a SENIOR ENGINEER reviewing a proposed change to how a function is
classified or organized in a handbook for an AI agent harness. Your job is to be
skeptical and find real concerns rooted in code behavior.

Focus on:
  - Does the proposed classification reflect what the function ACTUALLY does at the
    code level?
  - Are the line ranges / region boundaries syntactically and semantically sound?
  - Does the proposal misuse cross-cutting categories (e.g. labeling a function
    "crosscut-X3 logging" just because it calls self.logger)?
  - Are caller/callee relationships consistent with the proposed stage membership?

You may APPROVE, REVISE (with concrete concerns), or REJECT.

DECISION GUIDELINES
- APPROVE: the proposal is reasonable, even if minor wording could be tweaked. A
  correct-enough proposal is APPROVE, not REVISE. Be GENEROUS with APPROVE.
- REVISE: only when there is a SPECIFIC, ACTIONABLE flaw that materially affects
  correctness (wrong stage assignment, wrong region boundary, factual error about
  the code). Do NOT REVISE for style, phrasing, or "could be more thorough" reasons.
- REJECT: only when the proposal is fundamentally wrong and revision can't save it.

By default lean APPROVE. The pipeline must make progress.

OUTPUT FORMAT
Return ONLY a single JSON object wrapped in a ```json fenced block:

{
  "decision": "APPROVE" | "REVISE" | "REJECT",
  "concerns": ["<concern 1>", ...],
  "suggested_revision": { ... } OR null,
  "rationale": "<one sentence explanation>"
}

When decision == "APPROVE", concerns and suggested_revision may be empty.
When decision == "REVISE", concerns must be non-empty AND each concern must point
to a concrete, fixable flaw.
When decision == "REJECT", concerns must be non-empty; suggested_revision is null.
\end{lstlisting}

\subsubsection{Generating File Cards}

Used in \hblarge{} Phase~II to summarize each scanned file. In deep mode---where
the file is the handbook's L3 leaf---the model reads the whole file and writes the
card content, while the graph-derived function inventory, line numbers, and call
relations are supplied as fixed facts. The deep-mode prompt is:

\begin{lstlisting}[style=promptbox]
You are reading SOURCE FILES IN FULL and writing a plain-language, easy-to-follow
description of each, for a system handbook in which the FILE is the smallest unit
(its leaf node). The description you write IS the handbook's content for this file.

WHO YOU ARE WRITING FOR: a curious OUTSIDER -- someone intelligent but new to this
project and possibly not an expert programmer in this area. They should be able to
read your text and come away genuinely understanding what this file does and why it
matters, WITHOUT having to already know the codebase or the jargon.

HOW TO WRITE (this matters as much as the content):
- Use plain, everyday language. Prefer short, clear sentences over dense ones.
- Explain the WHY and the WHAT in human terms before any mechanism: what real
  problem does this file solve, and what would break without it?
- When you must use a technical term or acronym, explain it in plain words the
  first time. Never leave jargon unexplained.
- Use a brief everyday analogy when it genuinely makes something click.
- Do NOT dump implementation trivia. Favor the big picture and intuition. It is
  fine to name key types/functions, but always say what they're FOR in plain words.
- Stay accurate. Simplify, but never say something that is actually wrong.
- Avoid empty filler like "handles"/"manages related logic"; be specific.

Each file also comes with its FUNCTION LIST (qualname + line range), derived from
the call graph. The function inventory, line numbers, and call relations are
FACTS -- do NOT re-list them. Your job is the plain-language prose + an
easy-to-understand note per function (referenced by its exact qualname).

For EACH file return:
- "purpose": 1-2 plain sentences a newcomer can understand: what this file is for.
- "description": an accessible walkthrough (roughly 120-300 words): what problem
  this file solves and why it exists; what it does, step by step, in plain
  language; how its main pieces work together; and any surprising behavior.
- "functions": one entry per function in the file's function list, referenced by
  its exact "qualname", each with:
    - "purpose": in plain words, what this function does and why (1-3 sentences).
    - "data_flow": what goes IN -> what it does -> what comes OUT, told as a simple
      before/after story a non-expert can follow.
    - "relations": how it fits into the bigger flow -- who calls it and when, and
      what it hands off to, grounded in the provided calls/called-by facts.
- "role": EXACTLY one of: entrypoint, orchestration, domain_logic, io_transport,
  data_model, config, util, test, generated, other.
- "lifecycle": SHORT hint of when in the run this file is active (e.g. "startup",
  "config load", "main loop", "request handling", "teardown", "cross-cutting").

OUTPUT -- ONLY a JSON object in a ```json block:
{
  "purposes": [
    {"file": "<exact path>", "purpose": "...", "description": "...",
      "role": "<role>", "lifecycle": "...",
      "functions": [
        {"qualname": "<fn qualname>", "purpose": "...", "data_flow": "...",
          "relations": "..."}, ...]},
    ...
  ]
}
Return one entry per file given, using the exact file paths provided.
\end{lstlisting}

\subsubsection{Synthesizing the L1--L3 Document Tree}

The two modes synthesize the document tree differently. In \hbsmall{} mode, each
L1, L2, and L3 node is produced by a bounded actor--critic--reflexion loop: an
actor drafts the node from grounded inputs, a critic scores it against a
tier-specific rubric and returns actionable findings, and the actor revises until
the rubric passes, the score stops improving, or the generation budget is reached.
L1 and L2 use long, tier-specific writing guides (a whiteboard-tour style for L1;
a problem/main-flow/state-flow/hand-off structure for L2). In \hblarge{} mode, the
validated file cards from Appendix~\ref{sec:d-generation} already are the L3
content, and a single-pass rollup produces the L2 component overviews and the L1
system overview.

The L3 (function-as-leaf) entry is emitted against the output schema below,
grounded in the exact source range(s); a multi-region variant returns one gloss
per region with function-level design decisions and relations.

\Needspace{8\baselineskip}
\begin{lstlisting}[style=promptbox]
## Output format (single)

Return a JSON object inside a ```json fence:

```json
{
  "schema_version": 1,
  "type": "single",
  "locator_role": "<one-line role tag, 6-15 words>",
  "stage_context": "<2-4 sentences: what role this plays in the stage; when it is
                     invoked; how it relates to its siblings>",
  "synopsis": "<2-4 sentences: What + inputs + outputs + side effects>",
  "interface": {
    "signature": "<full signature, e.g. (self, environment) -> None>",
    "params": [{"name": "<param>", "type": "<type, or ? if unclear>",
                "role": "<one-line role / source>"}],
    "reads_state": ["<self._ attributes this function reads>"],
    "returns": "<what it returns; if None, state the real product (side effect)>",
    "side_effects": ["<self._ attributes written / external effects>"]
  },
  "execution_flow": [
    "<step 1: action + which args/state + what's produced>",
    "<step 2: ...>"
  ],
  "design_decisions": [
    "<decision 1: the trade-off point + why this choice + what an alternative costs>",
    "<decision 2: ...>"
  ],
  "relations": {
    "callers": ["<qualname or brief description>"],
    "core_callees": ["..."],
    "config_state_sources": ["..."],
    "results_to": ["..."],
    "siblings": [],
    "register_interactions": [
      {"action": "write|read|clear|reset", "register": "reg-<id>",
       "note": "<5-12 words: when / why>"}
    ]
  }
}
```

Count constraints:
  - interface.signature: required, copied accurately from the source.
  - interface params / reads_state / returns / side_effects: list them all from the
    source; empty array / "none" if there are none.
  - execution_flow: 2-8 steps; design_decisions: 1-5 items.
  - relations.{callers, core_callees, config_state_sources, results_to}: at least 4
    of these must be non-empty.
  - relations.register_interactions: list every register this function
    reads/writes/resets; empty array [] only if it interacts with none.
\end{lstlisting}

\subsection{BGPD Localization and Planning}
\label{sec:d-bgpd}

This is the planner prompt used under BGPD in the Handbook-Assisted arm: it
localizes by routing through the handbook and then verifying candidate edit sites
against the real source before emitting a self-contained edit plan. Per-harness
fields substituted at runtime are shown as \texttt{\{\{...\}\}} placeholders.

\begin{lstlisting}[style=promptbox]
You are a senior software engineer PLANNING a change to {{PROJECT_INTRO}}, on
behalf of a code reviewer.

You are given ONE natural-language change request. In this phase you produce a
precise, SELF-CONTAINED PLAN of the edits -- you do NOT edit any files. A separate
executor will apply your plan MECHANICALLY: for each edit it will substitute your
exact OLD text with your exact NEW text, WITHOUT re-reading the file. So your
plan's verbatim text must be byte-exact.

## Two artifacts, two distinct roles
- The handbook is a pure LOCATION INDEX of the harness, NOT a description of the
  code. Each function appears as a one-line locator: `<summary><b>Qualified.name</b>
  -- file:start-end . one-line role</summary>`, optionally followed by a Relations
  block (callers/callees/register read-write sites). A card has NO function body and
  NO source code. On top of that, `index.md` lists every stage with its function
  locators and `registers.md` lists every state variable with its exact read/write
  code sites. Use these to decide WHICH files, functions and sites are in scope --
  they surface scattered, non-obvious sites (mirror copies in the other
  parser/template, a register's every read/write, cross-subsystem touch points)
  that a plain text search can miss.
- The real source code is the GROUND TRUTH for WHAT to change. The handbook tells
  you the ADDRESS (file:start-end); the code at that address is the only reliable
  source of the actual structure. You MUST read the real source.

## How to plan -- ROUTE with the handbook, READ the real source, EMIT verbatim edits
1. Understand the request's true intent: the behavior delta, and the
   state/conditions/values it fixes.
2. Route with the handbook. Read its `SKILL.md`, then `index.md`, then only the
   `stages/<id>.md` chapters and `registers.md` entries your intent points to.
   Assemble the candidate set: every file + function + anchor the change must touch.
   Watch for scattered/mirror sites (a parser change usually has a twin in the OTHER
   parser and in both prompt templates; a state change fans out to every read site
   listed under that register).
3. Read the REAL source of every site you intend to edit with `read_file` on the
   actual code files. Confirm the exact body, control flow, conditions, and that the
   site does what the card implied.
4. For EACH edit, produce a self-contained EDIT BLOCK (format below) whose
   `old_string` is copy-pasted verbatim from the `read_file` output you just saw --
   never retyped from memory, never paraphrased. Match whitespace and indentation
   exactly, and include at least 3 lines of context BEFORE and AFTER the changed
   lines so the snippet is UNIQUE in the file.
5. A change can also silently break something the request never mentions; if you
   find such a coupled assumption, add an edit (or note it) accordingly.
6. Only include edits you are confident the request requires.

## EDIT BLOCK format (the executor applies these directly)
For every edit, output exactly:

### EDIT <n>
- file: `<path relative to the working dir, e.g. {{PATH_EXAMPLE}}>`
- where: `<{{WHERE_EXAMPLE}}>` -- why this change
```old
<EXACT current text, copied verbatim from read_file -- whitespace-perfect, unique>
```
```new
<the replacement text -- correct, idiomatic, the smallest change that realizes the intent>
```

Rules for the blocks (the executor trusts them blindly, so precision is on you):
- `old` MUST be byte-exact to the file's current content. If you are not certain it
  is verbatim, `read_file` that region again before writing the block.
- Keep each `old` the SMALLEST span that is still unique (1-8 lines typically); do
  not paste whole functions.
- SAME-FILE edits must NOT overlap. The executor applies your blocks in order
  against text that earlier blocks have ALREADY changed, and it does not re-read
  between them. So no block's `old` (including its context lines) may contain a line
  that another block in the same file changes -- otherwise the second match will be
  stale. If two changes are close enough that their context would overlap, MERGE
  them into ONE block covering both. Order same-file blocks top-to-bottom.
- For a brand-NEW file, use a single block with an empty ```old``` and the full
  content in ```new```, and say "(new file)" in `where`.
- Anchor on stable lines; never let `old` span a region you are unsure of.

When done, call `complete_task` with: a short prose summary of the edits, then ALL
the EDIT blocks, then the declarations JSON. Do NOT edit any files in this phase.

## Declarations (machine-readable -- the handbook-resync pipeline consumes this)
End with EXACTLY one ```json block declaring the change-set at FUNCTION
granularity, using {{QUALNAME_NOTE}}:

```json
{{DECL_JSON}}
```

- `will_modify` -- every EXISTING function whose implementation your edits change.
- `will_add`    -- every brand-new function introduced.
- `will_remove` -- every function deleted outright. A rename = remove(old)+add(new).
\end{lstlisting}

\subsection{LLM Judging}
\label{sec:d-judge}

Plan quality is scored in two steps. The judge first builds a leakage-safe answer
key from the request and pristine source, before seeing any plan, and then scores
each plan independently against that key.

\subsubsection{Building the Answer Key}

The judge builds a leakage-safe answer key from the request and pristine source
before seeing any plan (hint fields are stripped from the request). This is the
Terminus-2 judge; the Codex judge mirrors it, using \code{correct\_approach} in
place of \code{correct\_localization}.

\begin{lstlisting}[style=promptbox]
[SYSTEM]
You are an expert Python engineer building a grading ANSWER KEY for a code-change
task on the Terminus-2 Python agent harness.

You are NOT looking at any solution yet. Read the pristine source and independently
determine which concrete files, functions, methods, branches, templates, or state
variables a correct solution must localize before editing.

Output ONLY this JSON object:
{"required_sites":["<concrete file/function/branch/template/state site>"],
"correct_localization":"<2-4 sentences: where the planner should look and why>",
"pitfalls":["<easily missed mirror/fallback/cold/state/template sites>"]}

[USER]
## CHANGE REQUEST
{query}

## PRISTINE TERMINUS-2 SOURCE
{pristine}

Build the localization answer key now. Output the JSON object only.
\end{lstlisting}

\subsubsection{Scoring the Plan}

Each plan is scored independently against the answer key on a $0.0$--$5.0$ scale in
$0.5$ steps. The three dimensions are combined with weights $0.50$
(\code{localization}), $0.25$ (\code{scope\_bloat}), and $0.25$
(\code{reasoning}); \code{scope\_bloat} is reported as Scope Control, where a
higher value means a more focused plan.

\begin{lstlisting}[style=promptbox]
[SYSTEM]
You are an expert Python engineer GRADING the LOCALIZATION quality of one planner's
PLAN for a code-change task on the Terminus-2 Python agent harness.

You judge the PLAN ONLY: which files/functions/state/template sites it decides to
inspect or modify, and whether its reasoning is grounded. Do NOT grade the final
code diff or execution correctness. The executor is shared across arms and is out
of scope.

Grade these dimensions from 0.0 to 5.0 in 0.5 steps:
- localization : did the plan identify ALL required implementation sites in the
                 answer key, including mirror, fallback, cold-path, state-reset,
                 and template sites?
- scope_bloat  : did the plan avoid unrelated files/functions or overly broad
                 edits? 5 means tightly scoped; low means it wandered into
                 irrelevant areas.
- reasoning    : is the plan's localization reasoning grounded in the real code and
                 request, rather than vague, circular, or guessed?

Output ONLY this JSON object:
{"localization":X,"scope_bloat":X,"reasoning":X,
"justification":"<<=60 words: decisive localization hit/miss -- which sites found or missed>"}

[USER]
## CHANGE REQUEST
{query}

## ANSWER KEY
{answer_key}

## PRISTINE TERMINUS-2 SOURCE
{pristine}

## THIS ARM -- plan.md
{plan}

Grade this arm's PLAN localization against the answer key. Output the JSON object only.
\end{lstlisting}

\section{Illustrative Handbook-Assisted Planning Walkthrough}
\label{app:case-study}

This appendix walks through one real Handbook-Assisted run on Terminus-2 (case
\textbf{Q1}, DeepSeek-V4-Pro planner), reconstructed from its recorded trace. The
run makes concrete the two moves at the heart of BGPD: the planner first
\emph{routes} through the handbook to decide which code sites are in scope, and
then \emph{verifies} each candidate against the live source before writing a
plan. Every tool call, source locator, and edit shown below is taken from the
run; only whitespace is trimmed.

\begin{casebox}{Change request (Q1: triple completion confirmation)}
Right now the model only has to mark \code{task\_complete} once to reach the
``are you sure?'' step, then once more to get graded. I want a higher bar: the
model must mark completion \textbf{three consecutive times} before we actually
grade. The first two times just re-show the terminal state and ask ``are you sure
you want to submit?''.
\end{casebox}

\noindent The request is Query-type: it revises an existing control-flow rule but
names no file or function. The planner's first task is therefore one of
localization --- finding every site that implements the completion handshake.

\begin{casebox}{Phase 1 --- Route through the handbook}
Guided by the \code{SKILL.md} policy, the planner reads the handbook top-down,
narrowing the scope before opening any source:
\begin{enumerate}[leftmargin=1.4em,itemsep=1pt,topsep=2pt]
  \item \code{SKILL.md} $\rightarrow$ \code{overview.md} (L1) $\rightarrow$
  \code{index.md}: locate the stage that owns per-iteration continuation and
  termination.
  \item \code{stages/stage-4.8.md} (\emph{Completion Gate}) and
  \code{stages/stage-4.7.md} (\emph{Command Execute}): the handshake lives in the
  gate.
  \item \code{registers.md}: the gate is backed by the state register
  \code{reg-pending-completion}, whose read and write sites span the loop,
  initialization, and per-run reset.
\end{enumerate}
Two facts fall out of this. The change is confined to a single stage (Completion
Gate) and a single register; and the register's write sites reveal that
initialization and per-run reset must change together with the loop, not just the
obvious return branch.
\end{casebox}

\begin{casebox}{Phase 2 --- Verify against the live source}
Because the handbook is a pure location index and stores no source, the planner
now opens the real \code{terminus\_2.py} to confirm each locator. One
\code{search\_file\_content} for \code{\_pending\_completion} returns \textbf{seven
occurrences, all in \code{terminus\_2.py}}, so the change stays within one file.
Reading them shows a single Boolean flag that is set on the first completion mark,
tested before returning on the second, and cleared on any non-completion turn and
at the start of each run. This resolves the handshake to three concrete sites:
\begin{itemize}[leftmargin=1.4em,itemsep=1pt,topsep=2pt]
  \item \code{Terminus2.\_\_init\_\_} ($\sim$L292) --- initializes the flag.
  \item \code{Terminus2.\_reset\_per\_run\_state} ($\sim$L1574) --- clears it per run.
  \item \code{Terminus2.\_run\_agent\_loop} ($\sim$L1427--1440, L1552--1559) ---
  the observation branch and the completion gate that read and write it.
\end{itemize}
\end{casebox}

\paragraph{Edit plan.}
With the sites confirmed, the plan replaces the Boolean \code{\_pending\_completion}
with an integer counter \code{\_completion\_confirmations}: the loop increments it
on each completion mark, resets it on a non-completion turn, and returns for
grading only on the third consecutive mark. The change set is four \emph{modify}
actions, all in \code{terminus\_2.py} --- initialization, per-run reset, and the
loop's two branches. The completion gate itself is the representative edit:

\noindent\textbf{Before} --- two-mark Boolean handshake:
\begin{lstlisting}[style=codesnip]
if is_task_complete:
    if was_pending_completion:
        # second task_complete: confirmed, grade now
        return
    else:
        # first attempt: ask to confirm, continue
        prompt = observation
        continue
\end{lstlisting}
\noindent\textbf{After} --- three-mark counter:
\begin{lstlisting}[style=codesnip]
if is_task_complete:
    if was_completion_confirmations >= 2:
        # third consecutive task_complete: confirmed, grade now
        return
    else:
        # first or second attempt: ask to confirm, continue
        prompt = observation
        continue
\end{lstlisting}

\end{document}